\documentclass{article}

\usepackage{arxiv}

\usepackage[utf8]{inputenc} 
\usepackage[T1]{fontenc}    
\usepackage[hidelinks]{hyperref}       
\usepackage{url}            
\usepackage{booktabs}       
\usepackage{amsfonts}       
\usepackage{nicefrac}       
\usepackage{microtype}      
\usepackage{graphicx}
\usepackage{doi}

\usepackage{amsmath}
\usepackage{enumitem}
\usepackage{makecell}
\usepackage{multirow}
\usepackage{rotating}
\usepackage{tablefootnote}
\usepackage{imakeidx}
\usepackage{changepage}

\newenvironment{sidenote}
{\begin{adjustwidth}{2cm}{1cm}\footnotesize \textbf{Sidenote:\\}}{\end{adjustwidth}}

\newenvironment{definitions}
{\begin{adjustwidth}{2cm}{1cm}\footnotesize \textbf{Definitions:\\}}{\end{adjustwidth}}

\title{Knowledge Engineering for Wind Energy}

\author{ \href{https://orcid.org/0000-0002-6570-6966}{\includegraphics[scale=0.06]{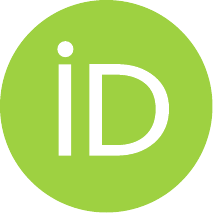}\hspace{1mm}Yuriy~Marykovskiy}, 
	\href{https://orcid.org/0000-0002-6436-8651}{\includegraphics[scale=0.06]{orcid.pdf}\hspace{1mm}Thomas~Clark}, 
	\href{https://orcid.org/0000-0002-5226-6105}{\includegraphics[scale=0.06]{orcid.pdf}\hspace{1mm}Justin~Day}, 
	\AND
	\href{https://orcid.org/0000-0002-1646-1691}{\includegraphics[scale=0.06]{orcid.pdf}\hspace{1mm}Marcus Wiens}, 
	Charles Henderson, 
	\href{https://orcid.org/0000-0002-1460-9808}{\includegraphics[scale=0.06]{orcid.pdf}\hspace{1mm}Julian Quick}, 
	  \href{https://orcid.org/0000-0001-8678-0965}{\includegraphics[scale=0.06]{orcid.pdf}\hspace{1mm}Imad Abdallah}, 
        \AND
	\href{https://orcid.org/0000-0003-4124-9040}{\includegraphics[scale=0.06]{orcid.pdf}\hspace{1mm}Anna~Maria Sempreviva}, 
	\href{https://orcid.org/0000-0002-0364-6945}{\includegraphics[scale=0.06]{orcid.pdf}\hspace{1mm}Jean-Paul Calbimonte}, 
	\href{https://orcid.org/0000-0002-6870-240X}{\includegraphics[scale=0.06]{orcid.pdf}\hspace{1mm}Eleni Chatzi}, 
        \href{https://orcid.org/0000-0002-7329-6434}{\includegraphics[scale=0.06]{orcid.pdf}\hspace{1mm}Sarah Barber}\\
}

\renewcommand{\shorttitle}{Knowledge Engineering for Wind Energy}

\hypersetup{
pdftitle={Knowledge Engineering for Wind Energy},
pdfsubject={AI},
pdfauthor={Marykovskiy et al.},
}

\begin{document}
\maketitle

\begin{abstract}
With the rapid evolution of the wind energy sector, there is an ever-increasing need to create value from the vast amounts of data made available both from within the domain, as well as from other sectors. 
This article addresses the challenges faced by wind energy domain experts in converting data into domain knowledge, connecting and integrating it with other sources of knowledge, and making it available for use in next generation artificially intelligent systems. 
To this end, this article highlights the role that knowledge engineering can play in the process of digital transformation of the wind energy sector. 
It presents the main concepts underpinning Knowledge-Based Systems and summarises previous work in the areas of knowledge engineering and knowledge representation in a manner that is relevant and accessible to domain experts.
A systematic analysis of the current state-of-the-art on knowledge engineering in the wind energy domain is performed, with available tools put into perspective by establishing the main domain actors and their needs and identifying key problematic areas. 
Finally, guidelines for further development and improvement are provided.
\end{abstract}


\section{Introduction}
\label{sec:introduction}
\subsection{Extracting value from data}
In the wind energy sector, it is becoming increasingly important to create value from data~\cite{veers2019grand}. To this end, vast amounts of data generated by various sources, including sensors and other monitoring systems, need to be effectively structured and represented in a way that can be easily understood and processed by both Artificial Intelligence (AI) systems and humans. 
The digitalisation of the wind energy sector is one of the key drivers for reducing costs and risks over the whole wind energy project life cycle \cite{Klonari2021WEDigitalisation}. The digitalisation process encompasses solutions such as digital twins, decision support systems and AI systems, some of which need to still be developed, in order to contribute to reducing operation and maintenance costs, for increasing the amount of energy delivered, as well as for maximising 
the efficiency of wind energy systems. 
In this context, the term \textit{Knowledge-Based Systems} (KBS) refers to AI systems that formalize knowledge as rules, logical expressions, and conceptualisations~\cite{akerkar2009knowledge,davis1986knowledge}. Such systems can be realised as AI-enabled digital twins or decision support systems that rely on databases of knowledge (also referred to as \textit{knowledge bases} or \textit{knowledge graphs}), which contain machine-readable facts, rules, and logics about a domain of interest, to assist with problem-solving and decision-making~\cite{hogan2021knowledge}.

\subsection{The need for managing data}
Currently, the stage for the digital transformation in wind energy is set by the democratisation of computing, technological maturity of AI systems, and the reduction in costs of data storage and sensing technologies. Along with this, a necessity to structure, organise, manage, and make use of substantial amounts of operational and synthetic data has emerged. However, it is often the case in industrial settings that data is not treated as an asset. 
Even though the importance of efficient data management has already been recognised by major stakeholders both in industry and academia~\cite{veers2019grand}, only a few organisations can afford to have a person dedicated to oversee data-related activities~\cite{Clifton2022}. This has left many domain experts one-on-one with the problems related to the actual, practical use of data~\cite{Barber2023usecasesWE}. The FAIR data approach, stating that data should be Findable, Accessible, Interoperable and Re-usable, introduced by \cite{Wilkinson2016}, provides general data management guiding principles. However, FAIR has mostly been applied in academic settings, and there is a disconnect between conceptual or descriptive guidelines and concrete implementations or defined prescriptions and practices. Several groups such as GO FAIR\footnote{\url{https://www.go-fair.org/}, Cited on 10.09.2023}, the Data Readiness Group\footnote{\url{https://datareadiness.eng.ox.ac.uk/ }, Cited on 15.09.2023}, and the Research Data Alliance\footnote{\url{https://www.rd-alliance.org/}, Cited on 16.09.2023} have emerged in recent years in an effort to provide practical implementation recommendations and solutions for increasing the FAIRness of data. Nevertheless, creating FAIR data frameworks still remains one of the major challenges in the digitalisation process~\cite{Clifton2022}.

\subsection{The challenge addressed in this paper}
Wind energy experts facing the challenge of managing their data will most likely find themselves overwhelmed by unfamiliar terms such as \textit{data schema}, \textit{relational data-model}, and \textit{metadata}. They may ask questions such as \textit{``What are the differences between a Structured Query Language (SQL) database, graph database, and an object store?''},  \textit{``Which one would fit best to my data types?''}  or \textit{``How do I publish my data on the web so that it conforms to the FAIR principles?''}. The same holds true during practical development of AI-enabled systems. In this context, a wind energy domain expert is increasingly expected to grasp concepts such as schema/ontology development, logic, and semantic networks, among others. Moreover, they often have to interact with a rather complex technology stack that includes data formats like Extensible Markup Language (XML), JavaScript Object Notation (JSON) or Semantic Web Technologies\footnote{\url{https://www.w3.org/standards/semanticweb/}} such as Resource Description Framework (RDF)~\cite{schreiber2014rdf}, SPARQL Protocol and RDF Query Language (SPARQL)~\cite{w3c2013sparql}, Web Ontology Language (OWL)~\cite{hitzler2009owl}, etc. 
These issues are particularly acute in the wind energy sector due to the fact that the industry is relatively new, the systems are highly multidisciplinary, and the relevant disciplines are currently highly siloed~\cite{Clifton2022}. Additionally, the modelling and measurement uncertainties are largely due to the high complexity of wind energy systems and difficulties in measuring operational data. As a result of all the factors mentioned above, wind energy data is often uncertain, incomplete, scarcely available, and context-lacking.
Recent efforts that have started to address the problem of data and knowledge management in the wind energy domain have not yet gained traction in the community. This is due to difficulties of cross-domain interactions, knowledge silos, lack of awareness among stakeholders, and other cultural and organisational factors \cite{Heidenreich2022knowledgeGenerationWE, Clifton2022, Kirkegaard2023}.
The need for holistic knowledge-based systems, however, is increasingly providing the necessary external pressure for the natural evolution and emergence of commonly accepted and adopted paradigms.

\subsection{Contribution of this paper}

To address the aforementioned challenges, this paper presents the main concepts and summarises previous work in the areas of knowledge engineering and knowledge representation in a manner that is relevant and accessible to wind energy domain experts. The insights presented in this article are not only beneficial for the wind energy sector, but also applicable to other domains undergoing digitalisation.   

The paper is structured as follows: Section~\ref{sec:KE_scope} presents the scope of knowledge engineering activities and the common roles in the overall context of digitalisation. Section~\ref{sec:KR_conceptual} presents a conceptual overview of the knowledge representation problem in general and introduces the basic concepts, or \textit{vocabulary}, of the knowledge engineering domain. Section~\ref{sec:KR_technologies} discusses practical technological implementations that enable the adoption of knowledge representation solutions, with a specific focus on Web technologies. Section~\ref{sec:KE_KBSystem} discusses how the knowledge representation technologies and knowledge engineering techniques enable the development of AI systems, in particular, AI-enabled digital twins. Section~\ref{sec:KE_in_WindEnergy} presents a systematic and methodological overview of the current initiatives by the wind energy community in the knowledge engineering domain. The overall state-of-the-art for knowledge engineering in the wind energy context is then discussed Section~\ref{sec:discussion}, followed by the concluding remarks in Section~\ref{sec:conclusions}.

\section{Knowledge engineering: scope and activities}
\label{sec:KE_scope}
%
Knowledge engineering refers to activities related to the development of AI systems capable of processing, interpreting, and performing logical operations on structured data~\cite{studer1998knowledge}. Knowledge representation refers to representing, or structuring, real world information in a way that renders this exploitable by AI systems. This involves choosing an appropriate representation language or formalism 
and determining how to map knowledge from the real world to the chosen representation.

Knowledge engineering activities often overlap with \textit{data management}. In particular, the creation of conceptual data models (also referred to as \textit{semantic data models}), which, conventionally, fall under the umbrella of data management, are also instrumental in the development of KBSs.
When designing a KBS, conceptual data models are used to represent the knowledge needed by the said system.
In a broader context, data management also involves activities related to the storage and maintenance of this knowledge. 
This includes defining how the knowledge is structured and stored, how it is accessed and updated, how its quality is ensured, and how it is integrated with other systems. Although data modelling activities, like knowledge representation, also involve the structuring of data, the focus is slightly different from that in knowledge engineering. Knowledge engineering is focused on capturing, representing knowledge, and logical reasoning and inference. Data management is focused on the overall process of collecting, storing, and using data within an organisation. As part of this process, conceptual data modelling is focused on the identification and organisation of key concepts and relationships. 
\begin{figure}[t]
    \centering
    \includegraphics[width=0.57\textwidth]{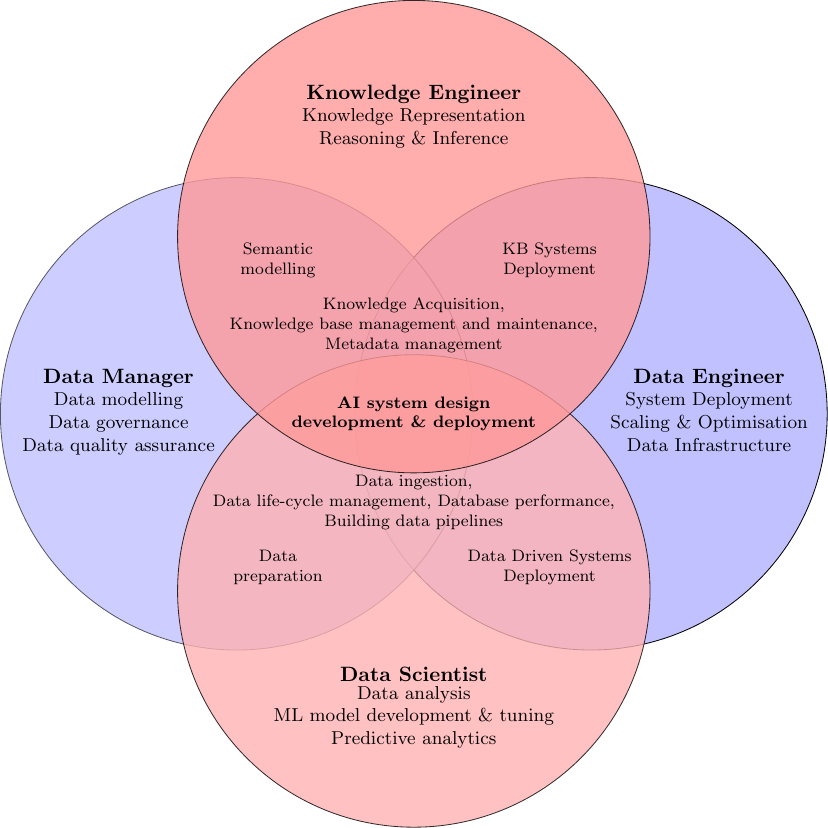}
    \caption{Typical roles and activities within the digitalisation process}
    \label{fig:ke-ecosystem}
\end{figure}

In Figure \ref{fig:ke-ecosystem} some typical roles and activities are presented. 
It is important to note that, in practice, it might not always be possible to clearly distinguish between these actors in a given organisation and there is no uniquely agreed upon classification. For example, roles like data modeller and/or database designer are often considered to reflect a narrower role of a data manager, as opposed to standalone positions. Terms like data steward are in some cases used interchangeably with data manager, while in others are used to denote more specific roles like the ones related to data governance. For the purpose of this paper, we distinguish the following roles: data manager, data engineer, data scientist, and knowledge engineer. A wind energy domain expert involved in the process of digitalisation is likely to interact with all these types of actors. This requires understanding of the jargon, workflows, and methodologies used in each respective domain. For this reason, some of the concepts from data science, data management, and data engineering are also discussed in the present work. However, given the vastness of these topics, this paper cannot be an exhaustive source on all these matters, but rather a simplified overview for a domain scientist of important concepts, directing the readers to the relevant works.

\section{Knowledge representation: conceptual overview} 
\label{sec:KR_conceptual}
%
In order to understand the practical application of knowledge engineering, it is important to be familiar with the concepts of knowledge representation and formal systems in general. The starting point of the discussion is human-centric. It revolves around the human perception, interpretation, and understanding of the world. The question of how humans model the world is an open-ended one, and, consequently, the question of how the knowledge should be represented and shared does not have a unique answer. 
This discussion about the nature and organisation of the world is the area of interest of a branch of philosophy called Ontology. In a knowledge engineering context, the underlying assumption is that the world, or domain of interest, consists of \textit{entities}, \textit{relationships}, and \textit{concepts}. 

Knowledge representation deals with the problem of capturing the meaning of facts (i.e. the aforementioned entities, relationships, and concepts) from a certain domain of interest  in a formal way as structured data. For example, consider the following \textit{informal} text:

\begin{quote}
\small
    An IET group at OST acquired an Aventa AV-7 wind turbine, located in Winterthur and has instrumented one of the blades with a novel pressure measurement system. The dataset produced by the system for the month of July is now available upon request. Additionally, during the measurement campaign wind turbine SCADA data was acquired and the inflow characteristics were measured with a LiDAR.
\end{quote}

A wind energy domain expert, especially a researcher with a background in experimental measurements, would not find it particularly difficult to understand and interpret this description. As a result, they can infer some additional information about the mentioned dataset. That is to say: the text provides contextual information about measurements, that a domain expert uses to assign meaning to particular data. However, there are several limitations to this representation of knowledge. First, this information is meant to be processed by a human (as opposed to some automated algorithm) with some command of the English language. Secondly, the reader must be a domain expert to infer the context and purpose of the text as well as resolve the inherent ambiguity of some of the statements. The underlying assumption is that a domain expert will rely on some informal logical framework~\cite{Groarke2022_informalLogic} and personal domain knowledge. 

\begin{sidenote}
     The term \textbf{formal} is mostly used as in ``formal system''. However, since AI systems are computational systems, i.e. machines, performing data manipulation based on a set of instructions (i.e. algorithms), in a knowledge engineering context, ``formal'' may also mean ``machine-interpretable''.  \textbf{Informal} is mostly used to denote something outside of such formal setting, for example assertions made using natural language~\cite{Johnson2002}.
     %
\end{sidenote}

In the example provided, some inferences are trivial for a human, while others are more complicated. For instance, it is clear that Aventa AV-7 is a type of a wind turbine. A domain expert will also have an understanding of what SCADA data may contain or the what LiDAR measurements may look like. Meanwhile, other information is not as straightforward. It is not clear what IET or OST is, especially if the text is presented by itself and not as a blog post on the university web page. Similarly, ``the month of July'' is not enough to pinpoint the time period without any additional context like the time and date when the text was written. Some information is fully absent, such as how the data is structured, what are the units used, whether the dataset is free and under what licence it is distributed.
This example brings up the concepts of semantics, pragmatics, context, metadata, language, logic, open world vs. closed world assumption, and ontology, which will be discussed below.

\subsection{Understanding representation: semantics, pragmatics, context, and metadata}
\label{ssec:semantics}
The process of understanding and interpreting a particular representation involves semantics, pragmatics and context. Semantics is the study of meaning in language, and is concerned with the relationship between words or symbols and their counterparts in the real world~\cite{Cann2009-CANSAI-2}. In the knowledge representation context, comprehensive semantics ensure that the terms used to describe data and information are unambiguous and clearly defined. Pragmatics, on the other hand, is concerned with the social and cultural factors that influence the use of language~\cite{andersen2011pragmatics}. In knowledge representation, pragmatics ensures that the meaning of a term or concept is understood in the appropriate social and cultural context. This can be particularly important when working with data or information from different disciplines or cultures.

In the realm of knowledge representation, context is pivotal as it impacts the interpretation of information, and ultimately its meaning. Context may encompass a variety of factors, including the data source, the conditions under which the data was collected, its intended use, or its relationship to other data. Metadata is data that provides this context as structured information about a data set. It is a form of formalised context, to make the representation not only human-interpretable but also machine-interpretable. 
For example, when publishing aerodynamic measurements and SCADA data on the Web, additional context can be provided by wind turbine characteristics. This contextual information can be expressed in a natural language (like English), in a form of technical specification sheets provided by the producer, or, ultimately, as some kind of formal representation. Such metadata would enable a data scientist to draw more meaningful conclusions while performing data analysis. For instance, the knowledge of the location along with historical weather data can provide understanding that a clustered group of measurements is due to an icing event or the specification of wind turbine status codes can link measurements to a certain wind turbine component failure.
The relevant question in this case is how to represent the knowledge about a particular wind turbine in a formal way. This is explored further as we discuss modelling languages, logic, and formalising representations. 

\subsection{Expressing representation: language}
In the example text given in the introduction to this section, the authors relied on the English language as a means of knowledge expression. Similarly, knowledge engineers and data managers rely on \textit{modelling languages} for knowledge representation and data structuring. On a fundamental level, the formal basis for modelling languages is provided by logic. Any conceptual model can be specified using some kind of \textit{logical language.}

\begin{definitions}
\textbf{Modelling languages} are formal languages that express information, knowledge or systems in a structure that is defined using a certain syntax (i.e. a consistent set of rules).
In this paper, we will discuss the most notable knowledge modelling languages in the context of knowledge engineering, computer sciences and information technologies. However, in the broader context of expressing knowledge, systems, and processes many other modelling language exist such as Unified Modeling Language (UML), Integration Definition (IDEF) languages, Petri Net, to name a few.\\
\textbf{Logic languages} are formal languages that provide a way to express logical statements and reason about them. Logic languages include syntax rules and a set of semantics that allow users to formally define and manipulate logical statements.amsmath
Examples of logic languages include predicate logic, description logics (DL), first-order logic (FOL), and fuzzy logic. 
\end{definitions}

For example, to represent the fact that Aventa AV-7 is a type of a wind turbine, it is possible to use first-order logic expressions:
\begin{align*}
\forall x:\, \mathrm{AventaAV7}(x) \rightarrow \mathrm{WindTurbine}(x)
\end{align*}
This statement can be read in English as ``For all \textit{x}, if \textit{x} is a AventaAV7, then \textit{x} is a wind turbine''. It is also possible to expresses similar semantics using description logics (DL) expressions:
\begin{align*}
& \mathrm{AventaAV7} \sqsubseteq \mathrm{WindTurbine}
\end{align*}
This statement can be read as ``All AventaAV7s are wind turbines''. These two statements in two different languages use different syntax to convey similar (but not exactly the same) semantics. 

The use of the logic languages in the context of information systems is rather impractical. As can be seen from the example above, the construction of a rather simple fact using FOL is often verbose and complex. This verbosity and complexity can result in misunderstandings, errors, and increased difficulty in managing and manipulating the data. Hence, in the domain of knowledge engineering and data modelling, specialised modelling languages are used, as discussed in Section \ref{ssec:ontology_languages}. 

\subsection{Interpreting representation: logic}
The choice of logic representation language depends on the desired semantics of the statements and on their complexity.
For example, a formal dataset description published on the Web that focuses only on the information presented, without a connection to other concepts or attributing additional semantics to the relationships with other entities does not require high expressivity. The metadata in such description might include fields for the specific turbine model (Aventa AV-7) and the location (Winterthur).
However, consider two statements about the OST-WindTurbine:
\begin{align*}
& \mathrm{locatedIn}(\text{OST-WindTurbine}, \mathrm{Winterthur}) \\
& \mathrm{locatedIn}(\text{OST-WindTurbine}, \mathrm{Switzerland})
\end{align*}
A more expressive language can define ``locatedIn'' to be a transitive relationship. If the system performing automated reasoning also has access to the fact that Winterthur is located in Switzerland, then a second statement can be automatically inferred without it being explicitly defined. 

Each logical language has its own way of achieving desired semantics. Consider making a following assertion: \textit{``There exists a wind turbine that has a rated power generation of more than ten megawatts of electricity''}. Using FOL, it can be formalised as follows:
\begin{align*}
\exists x(\mathrm{WindTurbine}(x) \land (\mathrm{ratedPower}(x)>10MW))
\end{align*}
In this statement, WindTurbine($x$) denotes that $x$ is a wind turbine, and $\mathrm{ratedPower}(x)>10MW$ denotes that $x$ is rated for more than ten megawatts of power generation. The semantics of this statement are following: if this assertion is true, it means there is a specific wind turbine $x$, for example the SG 11.0-200 DD model with ID0001, that meets the power requirement.
On the other hand, in DL it is possible to define a \textit{concept} of \textit{``a wind turbine that is rated for more than ten megawatts of power generation''} but not to make an \textit{assertion} about the existence of such a wind turbine.
\begin{align*}
 MegaWindTurbine \equiv (WindTurbine \sqcap \exists generates.MoreThan10MW)
\end{align*}
The statements about specific instance are done separately, for example:
\begin{align*}
    ID0001 :  SG11.0-200DD  \qquad SG11.0-200DD \sqsubseteq \mathrm{MegaWindTurbine}
\end{align*}
The difference in meaning is subtle, and is discussed more in Section \ref{ssec:OWA_CWA}, in which open world vs. closed world assumptions are discussed.

Even though, in general, most of the facts about the world can be described using FOL, it may be more practical to use languages based on DL for certain representations. For example, DL is especially powerful in situations where the goal is to represent knowledge in a structured and formalised way, such as in the creation of ontologies (see Section \ref{ssec:ontology} and Section~\ref{ssec:ontology_languages}).

\begin{sidenote} 
    It is difficult to define comprehensive semantics. This problem is not unique to the wind energy sector. Semantics for data modelling were explored deeply in the 1980s. Description Logic, in particular, frames the problem with RBox - ``roles box'', TBox - ``terminology box'' and ABox -  ``assertion box'' \cite{Rudolph2011FoundationsDL}. RBox captures interdependencies between the roles of the considered knowledge base. TBbox contains the definitions of the terminology or the concepts being used in the knowledge base (like a concept of the MegaWindTurbine). It consists of descriptions about what a particular concept or terminology means. The ABox, on the other hand, contains assertions or factual statements about instances of the concepts defined in the TBox (like the fact that ID0001 wind turbine is as an instance of SG11.0 - 200DD model of wind turbines). It essentially serves as a container for concrete examples of the definitions and relationships specified in the TBox and RBox.
\end{sidenote}

At the same time, both FOL or DL are impractical for representing certain facts and knowledge. For example, describing the observation of presence of cracks on a wind turbine blade along with their causes requires the use of additional knowledge representation methods such as physical models, simulation, and possibly probabilistic reasoning to accurately represent and reason about the state of the blade in question. The observation of cracks on a blade can be associated with uncertainties, as well as complex physical phenomena and causal relationships between load cycles, wind turbulence, and fatigue damage, which is difficult to model using solely first-order or description logic. While first-order logic can express relationships between objects and properties, it does not offer a good mechanism for expressing causality, observations of the states of the system, and related uncertainties. To express this statement more accurately, one would need to use a representation 
with different underlying logic (e.g. set theory), such as a causal Bayesian network, partially observable Markov decision process (POMDP) model or a dynamic systems model, which can capture complex interactions, states, transitions, and feedback loops.
\begin{definitions}
Formally \textbf{POMDP} is defined by a 7-tuple $(S,A,T,R,\Omega ,O,\gamma)$, where the items are sets of states $S$, actions $A$, conditional probabilities of moving from state to the other $T$, observations $\Omega$, conditional observation probabilities $O$, along with a reward function $R$ and a discount factor $\gamma$ 
\end{definitions}
For example, a specialist needs to decide the best time for a maintenance or inspection check on a set of wind turbine blades. The blades can be in one of two states: "Healthy" or "Damaged". However, these states are not directly observed. Instead, the specialist needs to rely on sensors and machine learning (ML) algorithms that provide indirect observations. Sometimes, these algorithms might give false alarms (false positive) or miss a fault (false negative). This problem can be represented as a POMDP, and solved for the optimal decision considering the uncertainties involved. 
\begin{itemize}
\item States $S: \{\mathrm{Healthy, Damaged}\}$
\item Actions $A: \{\mathrm{Maintenance, Inspection, NoAction}\}$
\item Observations $\Omega: \{\mathrm{CrackDetected, NoCrackDetected}\}$
\end{itemize}

With additional knowledge of the conditional probabilities involved (e.g. probability of detecting a crack with a ML algorithm, if the crack is present), and with defined reward function that reflects the benefit of performing action $a$ in state $s$ (e.g. inspecting a cracked blade might yield a high positive reward as it enables early intervention) the best course of action can be chosen. 
Evidently, a practical implementation of POMPD to solve the problem described is still challenging: an estimation of the conditional probabilities involved and a definition of the reward function is not straightforward. Moreover, the states of the blade are not discrete, but described by a wide spectrum of damage. Nevertheless, this example, however simplified, demonstrates how the choice of the formal representation is often dictated by the specific use case / problem. 


\subsection{Representation assumptions: open world vs. closed world}
\label{ssec:OWA_CWA}
The choice of logical language for knowledge representations can be influenced by the open world and closed world assumptions (OWA and CWA). 
In the CWA, it is assumed that everything not known to be true is false. This assumption is used in some logical languages such as FOL. In this context, the goal is to explicitly state all the necessary information about a domain and derive logical consequences based solely on this information. The CWA is useful in situations where the domain is well defined and the data is complete.
In contrast, the OWA states that everything not known to be true is simply unknown. This assumption is used in some logical languages such as DL. In this context, the objective is to define a set of axioms and a set of incomplete data. The logical consequences derived from these axioms and data are considered true until proven false. The OWA is useful in situations where the domain is complex, dynamic, and the data is incomplete.

Going back to the dataset publishing example, consider a person who would like to make a list of wind turbines that had some kind of measurement performed on them. Additionally, for the purpose of making a catalogue, this person would like to add information about owners / operators. 
One possible and intuitive approach is to make a table with two columns, "Turbine-ID" and "Owner-ID". While compiling the catalogue from the information available on the web, it is not unreasonable to assume that in some cases the owner will not be specified.
While it is clear that a conclusion within CWA such as "The wind turbine in question does not have an owner", is obviously far-fetched (for a human in the modern-day context), similar problems will naturally arise when dealing with information that can be legitimately ambiguous. For example: "Did the wind turbine in question also have strain gauges installed on it?" 
Bringing this example even further, consider the task of creating a common representation of a wind turbine. When describing a wind turbine it is reasonable to include the information about its gearbox. Direct drive wind turbines, however, eliminate the gearbox in their designs. In this case, the absence of data about gearboxes can also be ambiguous: the wind turbine in question might not have a gearbox, or the information was not available when the data was compiled.
Interestingly, this question is related to the admission of "Null" values in databases. It can be argued that if the information is incomplete, it should not be in the database in the first place. Of course,  when dealing with a complex domain such as wind energy and complex systems such as wind power plants, the idea of complete representation is rather ludicrous. 

\begin{sidenote}
    Consider the statement from before: $\exists x(WindTurbine(x) \land ratedPower(x) > 5 MW)$.
    To evaluate this statement, and obtain a True or False answer, we have to impose a certain restriction and assume complete knowledge of all the wind turbines and their power ratings in existence (or at least in the domain of interest). In an open world assumption, on the other  hand, the question of "existence" of a wind turbine that is rated for more that 10MW of power generation (just like the existence of pink elephants and unicorns) remains, well, open.   
\end{sidenote}

It is important to note that neither assumption is inherently superior—the choice between the two depends on the specific application and the nature of the data and knowledge being represented.
As mentioned above, some logical languages are designed to work with the CWA, while others are designed to work with the OWA. For example, adopting an OWA and FOL based representation would lead to undecidability. That is to say there is no algorithm to formally prove the "truthfulness" of the statements made using this logic. The FOL is the underlying logic for Structured Query Language (SQL) databases (See Section \ref{ssec:databases}). Thus these databases usually operate under CWA. The DL is the basis for Web Ontology Language (OWL), which is discussed in Section \ref{ssec:ontology_languages}.

\subsection{Formalising representations: ontology}
\label{ssec:ontology}

Ontology is a broadly used term that can take on different meanings. As mentioned in the introduction to this section, it can be used to denote a branch of philosophy. In the context of knowledge engineering and KBS, however, \textit{ontology} has been defined as ``explicit specification of a conceptualisation'' by Gruber \cite{Gruber1993}. Here, ``explicit'' means that the types of concepts used, and the constraints on their use, are explicitly defined. That is to say, each concept, attribute, relationship, and rule in the ontology is precisely articulated, often through formal semantics. This explicitness avoids ambiguity and fosters understanding, allowing the ontology to serve as a shared and common description of a domain that can be communicated across people and systems.
In addition, being explicit in an ontology also means that it is machine-interpretable. This is crucial for automated processing, reasoning, and interoperability in computer systems. With explicit ontologies, computers can process the semantic meaning of data, enabling more advanced and flexible uses of the data, such as inference and knowledge discovery.

``Conceptualisation,'' in Gruber's definition, refers to an abstract view or model of the world, i.e. the types of objects, concepts, and other entities that are assumed to exist in a domain of interest and their associated properties and relationships. An ontology, then, serves as a specific and concrete representation of that conceptualisation, allowing the underlying assumptions about the domain to be made explicit and facilitating their communication and processing. 

\begin{figure}
    \centering
    \includegraphics[width=0.95\textwidth]{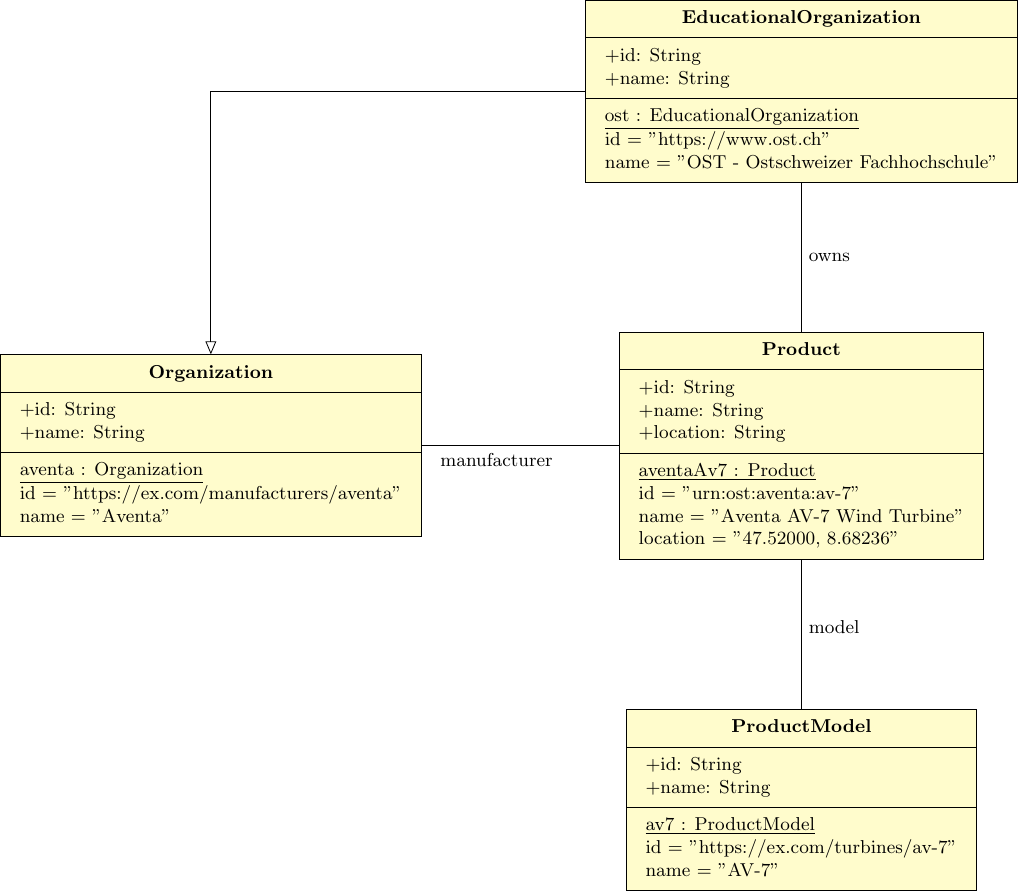}
    \caption{UML diagram representing a part of schema.org ontology as a labelled graph.}
    \label{fig:UML}
\end{figure}

\begin{sidenote}
    It is possible to think of an ontology as a directed labelled graph, where each "concept" (equivalently: "type" or "class") is represented as a node, and the edges represent the relationships (equivalently: "properties"). In fact, ontologies are often presented visually as a graph and expressed using graphical languages, as exemplified in Figure~\ref{fig:UML}.
\end{sidenote}

Given this general definition, \textit{Controlled Vocabularies}, \textit{Formal Taxonomies}, and \textit{Schemas} are also ontologies. Similarly, \textit{Conceptual (Semantic), Logical, and Physical Data Models} can also be thought of as ontologies.
 Generally, the difference in the use of these terms relates to the complexity (or ``expressiveness'') of the specification, with the term ``ontology'' typically being used to denote more expressive one \cite{Lassila2001}. In an effort to avoid ambiguity, some communities adopted an ``umbrella'' term \textbf{semantic artefact} \cite{le_franc_yann_2020semanticArtefact} to denote conceptualisations with various degree of expressiveness, and reserve the term \textit{ontology} only when referring to the most expressive conceptualisations.

 Another difference relates to the way and the context in which the specification is defined - a representation of a piece of (meta)data in a JSONSchema would be called a schema, whilst the same representation expressed in Web Ontology Language (OWL) would likely be referred to as an ontology. In the context of database design, a specification describing a structure of a database would more likely be referred to as database schema or logical data model or logical schema, rather than as an ontology \cite{Spyns2002DataModellingOntologyEngineering}. At the same time, an ontology that is populated with instances is often referred to as a \textit{knowledge graph} or a \textit{knowledge base} rather than as a database~\cite{heist2020knowledge}. This is done to distinguish these representations from relational databases (see Section \ref{ssec:databases}).
In this paper, the main focus is on the most expressive side of the ontology spectrum, as these can describe and formalise more complex relationships, which can facilitate the creation of the type of AI system needed for the digitalisation of wind energy, as discussed more in Section \ref{sec:KE_KBSystem}.

\begin{definitions}
A \textbf{controlled vocabulary} is a set of terms or phrases that have been pre-selected and authorised for use in a particular domain or context. It provides a standardised way of naming and describing concepts, which helps to improve consistency and accuracy in indexing, searching, and retrieval of information. In some contexts the term may be used to refer to a specific set of terms or concepts without any explicit relationships between them. However, in some fields the term may be used to denote a taxonomy. 

\noindent
 A \textbf{formal taxonomy} is a hierarchical classification conceptualisation that organises concepts or objects based on their relationships to one another. It is possible to view taxonomy as an ontology that includes only a subsumption relationship between classes. In practical terms it means that it possible to express a relationship of a type \texttt{isSublcassOf} (or equivalently \texttt{isA}) between the classes, thus modelling class hierarchy. Some taxonomies may, however, include other types of hierarchical relationships. For example, the Simple Knowledge Organisation System (SKOS) data model for taxonomy description defines semantic relationships of the type \texttt{skos:broader}, \texttt{skos:narrower}, \texttt{skos:related}, etc.

\noindent
A \textbf{schema} defines the relationships between different concepts and entities. In the knowledge engineering community, the term is used to refer to simple conceptualisations, like the ones written using less expressive knowledge representation languages such as RDF Schema\footnote{\url{https://www.w3.org/wiki/SchemaVsOntology} Cited on 01.10.2023}. In data management community, schema is usually used to indicate a blueprint or framework that defines the structure and content of a particular type of data or information (e.g JSON Schema for JSON data). It may specify the types of data elements that are allowed, their relationships to one another, and the rules for encoding or validating them.

 \noindent
 An \textbf{ontology} written in an expressive ontology language (See Section \ref{ssec:ontology_languages} for more discussion about ontology languages.), can represent rich, complex knowledge about concepts and their inter-relationships. For example, ontologies expressed with OWL-DL can define transitive, inverse, reflexive, and irreflexive relationships, impose cardinality restrictions etc.  
\end{definitions}

Using the dataset publishing example, it is possible to demonstrate how semantic expressiveness increases when moving from controlled vocabulary to ontology.
Controlled vocabulary can include concepts (equivalently: "terms") like "wind turbine", "pressure measurement system", "SCADA", and "LiDAR." Using a controlled vocabulary, it is possible to identify and label these key concepts in the paragraph, but there is no explicit representation of the relationships between these concepts or their properties. A taxonomy can include subsumption relationships between the concepts in the paragraph. For example, "LiDAR," and "pressure measurement system" can be subsumed by "measurement system." It is also possible to include other relationships, such as part-of relationships between the wind turbine and its blades, or between the measurement system and the turbine. A schema can include additional information about the properties of the concepts and their relationships. For example, it can specify the expected attributes of the wind turbine, such as its manufacturer, capacity, and location. Moreover, a schema may provide constraints that specify the relationships between these concepts: "the wind turbine's location is specified by a pair of latitude and longitude coordinates." Finally, an ontology can further specify the meaning and relationships of the concepts in a formal and machine-interpretable way. 
An aligned ontology will enable knowledge to be connected from different domains. For instance, the use of an ontology of geographical names can help to connect wind turbine data with specific locations and regions. Moreover, a formal ontology enables automated reasoning and inference of the concepts and their relationships, thus new knowledge can be derived from the information provided. This can form a base for KBS such as digital twins or decision support systems, which are becoming increasingly important in wind energy in order to reduce costs and increase deployment. 

In addition to semantic expressiveness, ontologies can be differentiated based on their scope or level of generality. Here it is common to distinguish top-level, domain, application and task ontologies \cite{Guarino1998FormalOntology}.
 \begin{definitions}
     \textbf{Top-level ontologies} provide a broad framework for organising concepts and relationships that are applicable across multiple domains or applications. These are useful for inter-domain knowledge exchange.\\
    \textbf{Domain ontologies} are aimed to capture domain-specific knowledge. They contain concepts and relationships that are relevant to a specific domain such as medicine or engineering. Adoption of these ontologies can ensure that the terminology used within a particular domain is consistent and clear.\\
     \textbf{Application ontologies} are designed to support a specific software application or system. These ontologies provide a more detailed and specialised vocabulary that is tailored to the needs of the application.\\
     \textbf{Task ontologies} are focused on the specific tasks or activities that need to be performed within a particular domain or application.
 \end{definitions}
 
Developing, publishing and using ontologies needs an effective collaboration among different actors including the ones mentioned in Section~\ref{sec:KE_scope}, but also domain experts, stakeholders, and target users, each with specific competencies and interests. 
Ontology Development 101 \cite{Noy2001} is a good starting point to familiarise oneself with the concept of ontologies, terminology used and development methods.
Another ontology development methodology particularly well suited for application and task ontologies and knowledge base development was proposed by \cite{DeNicola2005}. This method focuses on collaboration between domain experts and knowledge engineers during ontology development.

\subsection{Common representation: standard}
An ontology that is accepted and enforced by a certain community can be included in a standard. In the context of knowledge representation, a standard is a set of guidelines or specifications that prescribe how to represent and organise knowledge in a consistent and interoperable way. Standards ensure that knowledge representations can be shared, reused, and understood by different systems and applications, regardless of their implementation or environment. For example, the Dublin Core metadata schema, which defines essential metadata elements for the web publishing task (e.g creator, publisher, abstract, etc.), has been formally standardised as ISO 15836. 

Standard conceptualisations allow for standard data generation and transformation procedures. 
A central organisation publishes a comprehensive set of standard semantic artefacts, which can be updated based on community feedback, though any revision process will inherently be slow-moving and filled with compromises. It can be expected that organisations will develop and publish their own terminologies, schemas, or ontologies based on their specific needs and use cases. 

The development of standards can be affected by whether a OWA or CWA paradigm is adopted by the standard designers. 
An OWA standard semantic framework allows for modularity between different conceptualisations, where an ecosystem of different semantic artefacts can develop \cite{villegas-etal-2014-metadata,okgoogle}. This system allows anybody to define or iterate on an ontology or its sub-elements. This OWA framework is particularly useful for developing big-data insights \cite{rogushina2020semantic}. 
As an example, turbine Supervisory Control and Data Acquisition (SCADA) data is complex, involving hundreds of thousands of data variables, where the same metadata vocabulary can have different meanings between turbine models.
Most of these terms are defined in IEC 61400-25 standard. However, this standard does not include a machine-readable formalisation as part of its specifications, which would be needed for the digitalisation process (see Section \ref{sec:KE_in_WindEnergy}). 
SCADA ontology conforming to an OWA standard semantic framework would allow for standard data transformation procedures while maintaining model-specific semantic heterogeneity. Examples of data transformations standards include fault codes, power curve measurement, and damage estimation.

\section{Knowledge representation: technologies}
\label{sec:KR_technologies}
%
In this section, we discuss the practical aspects of knowledge representation. We will explore various technologies that have been developed to implement the theoretical concepts of knowledge representation we've discussed so far. This includes Semantic Web's vision for web of data and how it interrelates with FAIR principles, Resource Description Framework as way to to express graph-based data model, and the use of ontology and schema languages for expressing knowledge structures.

\subsection{Semantic Web}
\label{ssec:semantic_web}

\begin{figure}[t]
    \centering
    \includegraphics[width=0.9\textwidth]{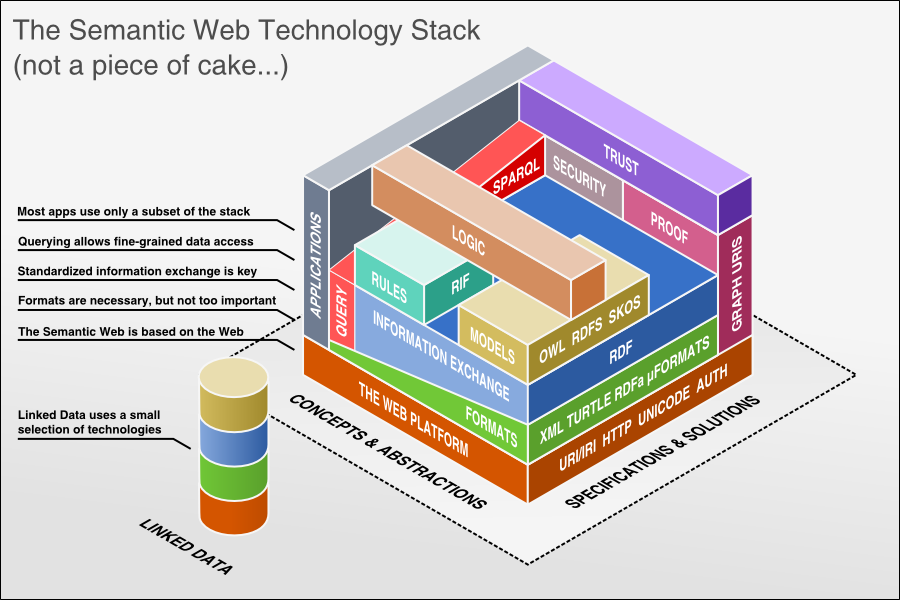}
    \caption{Semantic web stack \cite{Nowack2009}}
    \label{fig:semantic_web}
\end{figure}

The Semantic Web\footnote{https://www.w3.org/standards/semanticweb/} is an ambitious extension of the world wide web proposed by the World Wide Web Consortium (W3C) that seeks to create a "web of data" to make data more machine-readable and interoperable. The core idea of the Semantic Web is made possible through technologies discussed further below, such as the Resource Description Framework (RDF) and the Web Ontology Language (OWL), which allow data to be annotated and related in a machine-understandable way. At the same time conceptualisations and abstractions form the foundation of the Semantic web technology stack (Figure \ref{fig:semantic_web}). 

The web of data vision carries a wealth of practical advantages and has shown early dividends across various industries \cite{Noy2019industryKnowledgeGraphs}. 
For instance, e-commerce industry provides a valuable precedent for how the Semantic Web can deliver tangible benefits. Online retailers like Amazon and eBay use structured data to enrich product descriptions, enhancing product discoverability and improving customer experience. At the same time online advisement companies like Google rely on structured metadata and microdata\footnote{https://www.w3.org/TR/2021/NOTE-microdata-20210128/} for Search Engine Optimisation (SEO) by including it in their knowledge bases \footnote{\url{https://www.blog.google/products/search/introducing-knowledge-graph-things-not/}, Cited on 10.09.2023}. This structured data also allows for better integration with suppliers and logistics providers, creating a more seamless and efficient e-commerce ecosystem.

The web of data is poised to become transformative for the wind energy sector as well, helping to address key challenges around data use. Some of the significant benefits of the Semantic Web include:

\begin{itemize}
    \item \textit{Intelligent data discovery:} Semantic Web improves data discoverability by enabling search engines and applications to understand the context, content, and relationships of data. This can speed up data-driven investigations, like root cause analysis of turbine faults, by helping engineers quickly find relevant data and information.
    \item \textit{Data interoperability:} The Semantic Web allows for seamless integration of different data formats and sources. This becomes particularly beneficial in the context of the wind energy industry where heterogeneous data - ranging from wind speed measurements and power output statistics to maintenance records and weather forecasts - need to be integrated and analysed for effective decision-making. By structuring and interlinking data on wind turbines, weather conditions, maintenance activities, and grid demand, wind farm operators can create a rich, machine-readable data environment. This can enable intelligent applications, like AI-based predictive maintenance systems and decision-support, enhanced operational efficiency, and ultimately, decreased levelised cost of energy.
    Additionally, increased data interoperability facilitates data sharing and collaboration within the wind energy community. Development of shared ontologies for wind energy data can improve data exchange between different stakeholders - from wind farm operators and maintenance providers to equipment manufacturers and researchers. Such a collaborative approach could accelerate innovation and efficiency gains across the industry.
    \item \textit{Automation and AI readiness:} The machine-readable nature of Semantic Web data lays the groundwork for automation and AI applications. For the wind energy industry, this means the potential for advanced analytics, predictive maintenance, and automated optimisation of wind farm operations with AI augmented systems such a digital twins.
    \item \textit{Data reusability:} Semantic Web encourages the use of standardised schemas and ontologies, making data readily reusable across different contexts and applications. In the wind energy industry, this can facilitate cross-project and cross-site analytics, increasing the confidence in the analysis results and enhancing the understanding of wind turbine performance and reliability.
\end{itemize}

\paragraph{Interrelation with FAIR principles.}
A reader might notice a strong similarity between the benefits of the Semantic web and FAIR (Findable, Accessible, Interoperable, Reusable) principles, proposed by Wilkinson et al. \cite{Wilkinson2016}. While the discussion of FAIR principles falls more into the data management domain, the Semantic Web's vision, and specifically the concept of Linked Data (LD), \cite{Heath2011ld} is intertwined with the ultimate goals of FAIR approach.
According to W3C themselves, "Linked Data lies at the heart of what Semantic Web is all about: large scale integration of, and reasoning on, data on the Web." At the same time, technologies that enable LD also enable data FAIRness. In fact, the practical recommendations for increasing data FAIRness, such as publishing structured metadata on the web, refer to Semantic Web Technologies and LD \cite{wu2021_structuredMetadata}. For more discussion about how LD can enable FAIR data see Appendix \ref{app:fair_ld}.

While there is a significant overlap between the LD and FAIR principles in terms of their instrumental values, the fulfilment of one set of these principles does not generally imply the other. In fact, FAIR principles are descriptive in nature, and are technology independent. Moreover, while LD focuses on interoperability aspect and data openness, FAIR data principles are not restricted to open data. Additionally, FAIR principles introduce requirements of metadata persistence and adherence to community standards. 
To illustrate the difference in the two perspectives, we can consider how LD and FAIR data are evaluated. A common way to evaluate LD is the "5 star linked data" specification\footnote{https://www.w3.org/community/webize/2014/01/17/what-is-5-star-linked-data/}. This concept relies heavily on the use of RDF and other Semantic Web technologies. On the other hand, a structured approach to assess the FAIRness of data was proposed by RDA with their Data Maturity Model \cite{RDA2020DataMaturity}. This model seeks to create a standard understanding of FAIR principles across diverse stakeholder groups. However, it does not dictate the exact means of evaluation nor the specific technical solutions. Instead, it offers a degree of flexibility while assessing data FAIRness. This is indicative of the model's recognition of the diverse contexts in which data can exist and the different standards that may apply in different fields or sectors.

\begin{table}[t]
\centering
\begin{tabular}{p{2cm} p{5cm} p{5cm}}
\hline
\textbf{Technology} & \textbf{What is it} & \textbf{Most common use case} \\
\hline
RDFS & Data modelling language, for definition of a basic ontology & Creation and publication of ontologies and knowledge bases \\
& &\\
OWL & An ontology language built on RDF and RDFS allowing definition of more expressive ontologies (e.g. transitivity, reflexivity) & Creation and publication of ontologies \\
& &\\
SHACL & A schema language for describing RDF data & Validation of data in an RDF graph \\
& &\\
JSON-LD & A syntax for serialization of ontologies and linked data ("Lightweight Linked Data Format") & Adding context to properties within JSON data in order to attach semantic meaning from a linked ontology \\
& &\\
JSONSchema & A schema language for describing JSON (or JSON-serialisable) data & Communicating requirements and validating data at boundaries between applications/services (like APIs)/organisations \\
& &\\
YamlSchema & A schema language for description of YAML data & Linting and validating text-based configuration files \\
& &\\
XMLSchema & A schema language & Validating data at boundaries between applications/services (e.g. APIs), typically for legacy systems \\
& &\\
AvroSchema & A schema language for use with the AVRO serialization utility & Validating event data in high data-rate event-driven systems (e.g. usually high-volumes of small events with low complexity in data structure) \\
& &\\
HDF5Schema & A schema language for describing HDF5 (or HDF5-serialisable) data & Validating data in saved HDF5 data files (frequently used for scientific applications) \\
& &\\
Protobuf & A mechanism for serializing typed and structured data & Validating event data in high data-rate event-driven systems (e.g. usually high-volumes of small events with low complexity in data structure) \\
\hline
\end{tabular}
\caption{Overview of knowledge representation languages}
\label{tab:kr_technologies}
\end{table}

\subsection{Resource Description Framework}
\label{ssec:rdf}
In the context of knowledge engineering, the key technological foundation for ontological representation and information exchange is provided by the Resource Description Framework (RDF).
RDF is a W3C standard for representing knowledge in the form of a graph-based data model. It was initially designed as a metadata model for describing resources on the web. RDF serves as one of  the fundamental layers in the Semantic Web technology stack. In RDF, data is represented as triples, consisting of subject-predicate-object expressions. The subject is a resource, typically identified by a Uniform Resource Identifier (URI), representing the entity being described or related to another entity. The predicate represents a relationship between the subject and the object, typically identified by a URI as well. The object can be either another resource or a literal value. These triples form a directed graph that can be queried and reasoned about using various technologies such as SPARQL, a query language designed for RDF, and RDFS or OWL ontology languages (discussed further below in Section \ref{ssec:ontology_languages}) built on top of RDF.

An example of RDF statement using TURTLE (See  Appendix \ref{app:serialisation} for more information about serialisation formats) syntax:
\begin{verbatim}
    @prefix ex: <http://example.com/resource/> .
    @prefix rdf: <http://www.w3.org/1999/02/22-rdf-syntax-ns#> .
    
    ex:Aventa_AV-7 rdf:type ex:WindTurbine . 
\end{verbatim}
Here, Aventa AV-7 (subject) is linked to the concept of a Wind Turbine (object), by a "type" (predicate) relationship.
As discussed before, an ontology can be visualised as a labelled graph. RDF triplets is a natural way to describe a graph with Subject as a starting node, predicate indicating a label of the edge and the object as a target node.

\subsection{Ontology and schema languages}
\label{ssec:ontology_languages}
Any ontology or schema must be \textit{expressed} using a language. A Schema (or Ontology) Language is a combination of syntax and semantics (particular to each language) allowing the user to express the structure and content of data. A variety of languages exist to do this. Such languages can be more or less "feature-complete", i.e. their ability to express complex relations and semantics vary.

Generally, Ontology Languages are more oriented (in terms of their features and abilities) toward OWA data representation and relation, whilst Schema Languages are more oriented toward defining and validating CWA data structures. These two schools of thought are converging as the languages themselves evolve. For example, JSONSchema (which emerged for validation of closed data coming through web APIs) is increasingly moving toward a full ontological language \cite{Angele2021}, while Shapes Constraint Language (SHACL) provides closed world validation (in the manner of JSONSchema) on open graphs described by RDF.
\begin{sidenote}
   The language doesn't have to be a text-based language. For example, IDEF5 is a graphical language that can be used to express an ontology.
\end{sidenote}
A variety of schema languages were reviewed for the purpose of describing CWA data \cite{Clark2022}. 
The summary of some commonly adopted ontology/schema languages and other data representation technologies is presented in Table \ref{tab:kr_technologies}. A more detailed description and the examples of simple statements made using these languages can be found in Appendix~\ref{app:ontology_languages}.


\section{Knowledge engineering: knowledge-based systems}
\label{sec:KE_KBSystem}
%
Knowledge-based systems (KBS) are a class of intelligent systems that utilise knowledge engineering techniques to capture, represent, store, and manipulate domain-specific knowledge to solve complex problems, support decision-making, and enable advanced applications. 
This section explores how the next generation of AI systems, such as Cognitive Digital Twins (CDT) \cite{Zheng2021} or Autonomous-Management Digital Twins \cite{Wagg2020} can combine recent developments in machine learning (ML), uncertainty quantification (UQ), verification and validation (V\&V), Bayesian approaches and Decision Support Systems (DSS) with classical rule-based KBS. These hybrid systems are enabled through knowledge integration and interoperability as we discuss in Section \ref{ssec:knowledge_integration_interoperability}. Additionally we touch upon data management and data engineering aspects of creation of such systems in Section \ref{ssec:databases}.

\subsection{Digital twins}
\label{ssec:KBS_DT}
The digital twin (DT) conceptual model was initially introduced in the context of product life-cycle management by Grieves \cite{Grieves2002} and later adopted for a wide range of applications in various domains, including wind energy. The basis of DT model are the concepts of "duality" and "strong similarity" between the physical world and its digital representation  \cite{Grieves2022}. Practical manifestations of DT instances come in a variety of types, depending on the actual realisation of the digital object and the extent to which the "strong similarity" is achieved. Recently, several attempts at classification of the DT types have been made \cite{VanDerValk2021, Pronost2021, Uhlenkamp2022, Marykovskiy2022taxonomyDT}.

\begin{sidenote}
    In their joint position paper, the American Institute of Aeronautics and Astronautics and the Aerospace Industries Association proposed a general definition of a DT as a "virtual representation of a connected physical asset". Moreover, examples and added value of 17 different DT types are proposed. \cite{aiaa2021}
\end{sidenote}
As digital representation is at the core of the DT concept, knowledge representation and knowledge engineering methods can be, and often are, employed in the development of DT instances.
Such DTs, can also employ data science methods~\cite{Ding2019dataScienceWE} and DSS~\cite{Seyr2019dssReview} to offer advanced functionalities including integration of heterogeneous data sources, prediction of unmeasured and future quantities based on historical data, and capability to produce actionable insights from updatable models.
%
%

In the wind energy domain, DTs can be implemented at various system levels (components, assemblies, wind turbines, wind farms, and grid) and throughout the asset's life cycle starting from the design phase and ending with the decommissioning.
Data integration, or ontologies on a higher level, provide the backbone for the functional capabilities.
Heterogeneous interfaces of single systems can be connected with others, by describing the system with a knowledge graph.
Thereby, the orchestration of the interactions between subsystems and processes is enabled~\cite{Wagg2020}.
Semantic technology enables the verification of existing metadata, knowledge inference, and the creation of new knowledge via rule-based reasoners, thus providing cognitive capabilities for CDT-type systems~\cite{Zheng2021, Arista2023AircraftDesignCDT}.
Additionally, ontologies can be used to describe model interfaces for simulations used in digital twins.
In this case, the structure and variables of model inputs and outputs are described and can be utilised in the automated setup of a modular model~\cite{Wiens2021}. 

Knowledge engineering is crucial in developing digital twins as it integrates heterogeneous data, automates data management and data science workflows, and facilitates connections with other digital twins or models in larger systems. For DTs which include DSS, knowledge engineering provides the ability to perform complex queries, as well as reasoning and inference capabilities. Overall, knowledge engineering methods enhance the functionality and effectiveness of digital twins.

\subsection{Knowledge integration and interoperability}
\label{ssec:knowledge_integration_interoperability}
Knowledge integration and interoperability lies at the core of knowledge engineering.
\textit{Ontology-based data integration (OBDI)} has emerged as a powerful solution to consolidate and interoperate heterogeneous data sources, utilising ontologies as shared (or aligned), semantic schemas \cite{DeGiacomo2018OBDI}. Through the use of ontologies, OBDI enables the harmonisation of diverse data sources into a coherent, query-able whole, promoting knowledge discovery and inference across systems that may otherwise remain isolated. In the wind energy sector, OBDI could integrate disparate data sources such as weather forecasts, energy production logs, and maintenance records, promoting a comprehensive, multi-perspective analysis of wind turbine performance, reliability, and optimisation.

\textit{Ontology evaluation and alignment} are crucial for interoperability and OBDI. Ontology evaluation ensures suitability and quality of a given knowledge base. Methods for ontology evaluation may differ from one context to another. The author of \cite{Vrandecic2009OntologyEvaluation} proposes assessing the quality of an ontology by evaluating such properties as accuracy, adaptability, clarity, completeness, computational efficiency, conciseness, constituency, and organisational fitness. Ontology alignment identifies semantically equivalent entities from different ontologies, enabling the harmonisation of heterogeneous data sources. 
In practice, this can be implemented by connecting different concepts using OWL \texttt{owl:sameAs} or SKOS \texttt{skos:exactMatch} relations.
This can significantly benefit the wind energy industry by allowing disparate systems and databases to interact and exchange information seamlessly, promoting a more efficient and effective operational workflow. For example, different organisations perform reliability and failure analysis of using their own taxonomies of wind turbine parts. Aligning these taxonomies between themselves not only allows a more comprehensive analysis, but also significantly increases the amount of available data, resulting in higher confidence in analysis results.

\textit{Ontology reuse} is another important aspect of knowledge integration. Reusing existing ontologies can reduce the effort and complexity involved in developing new ontologies from scratch and promote interoperability by using shared semantic artefacts. 
An important tool for ontology reuse are ontology hosting services. The hosting and sharing of ontologies requires the use of platforms and repositories allowing the discovery, search, versioning, and interconnection of the semantic models. While ontology hosting has been initially performed for specific communities and domains, there are several common functionalities (search, identification, alignment, annotation, etc.) that are orthogonal to domain-specific aspects. An example of such application is the OntoPortal Alliance, a consortium constituted of multiple research institutions dedicated to the development and maintenance of the OntoPortal platform \cite{Graybeal2019AdoptionOB}, available as open-source code. Based on this common platform, different instances of the portal are made available to specific communities, as for example BioPortal \cite{Noy2009bioportal}, AgroPortal \cite{Jonquet2018agroportal}, EcoPortal \cite{Kechagioglou2021EcoPortalAE}, etc. Compared to other platforms and initiatives for ontology hosting, the OntoPortal platform provides not only the most comprehensive set of features, but also the widest adoption in different domains \cite{jonquet2023ontoportal}. We discuss possibility of ontology reuse in wind energy domain in Section \ref{sec:KE_in_WindEnergy}.


\subsection{Data storage and management for knowledge-based systems}
\label{ssec:databases}
The topic of data storage and management for KBS is where knowledge engineering overlaps heavily with the data management and data engineering domains.
In terms of data management, the relational model based on FOL \cite{Codd1990}, usually in the form of SQL databases, has been widely adopted since the early 1990s across all industries as a solution for creating and managing structured data. In recent years, non-relational systems such as Not Only SQL (NoSQL) databases have gained popularity, as they provide for more flexible database expansion, allow for multiple data structures, and offer better performance when scaling up to deal with large data sets \cite{Lourenco2015}. In the context of knowledge engineering, triple stores (such as Ontotext\footnote{https://www.ontotext.com}) and graph databases (such as Neo4j\footnote{https://neo4j.com/}) are well suited to provide a technological foundation for the development of Semantic Web applications \cite{Soussi2019}. However, before selecting a database for storing and managing ontology-based data, several consideration should be taken into account, as described in this section.

\textit{Relational and SQL databases} excel at organising data in a structured, tabular format. They are particularly powerful when dealing with large amounts of structured data that needs to be queried with complex logic, given their ability to perform reliable and robust transactions \cite{Haerder1983ACID}. SQL databases can be very efficient for look-ups and queries that involve tabular type data.
Nevertheless, fitting ontology-based data, which are more graph-like in nature, into the format of an SQL database can pose significant challenges.
As discussed before, the fundamental data model for ontology is graph-based, whereas a tabular data structure is typically relational. These two differing structures often do not align seamlessly, leading to issues in data management.
The term \textit{impedance-mismatch} is used to denote the issues that surface when a system tries to transform one type of data structure into another. Specifically, when data is mapped from a graph-like or an object-oriented model to a relational model, a mismatch arises due to the structural differences between these representations. Over time, a variety of strategies, often referred to as 'object-relational mapping' methods, have been developed to address this mismatch. 
These methods focus on transitioning data from object-oriented models (based on classes and objects) into a format suitable for storage in relational databases (based on tables and relations).

Regardless of these developments, the fundamental divide in modelling approaches remains. Ontology-based modelling focuses on concepts or objects and describing the relationships between these concepts. While SQL databases have introduced some object-oriented features, they haven't yet introduced rich modelling semantics that are seen in ontological approaches. 
Hence, in cases when data has complex relationships or when the relationships themselves are inherently valuable, NoSQL graph databases offer more efficiency in terms of query speeds.

\textit{Object stores and NoSQL databases} were developed as solutions to certain limitations that traditional SQL databases had, particularly in two aspects: handling larger amounts of data and dealing with a variety of data types. 

To understand the first aspect, it is important to understand the notion of "scaling". In simple terms, "scaling" refers to the increasing capacity to handle more data or requests. There are two main ways of doing this: "scaling up" (also known as vertical scaling) and "scaling out" (also known as horizontal scaling). Scaling up refers to improving the capacity of a single server, such as by adding more memory or a faster processor. However, there are physical limitations to the extent by which a single server can be upgraded. On the other hand, scaling out involves adding more servers to a system and distributing the data and workload among them. This can provide greater increases in capacity, and allows for more flexibility and resilience because if one server fails, others can take over its workload.

The second aspect, data variety, refers to the shift from storing data in tables (as SQL databases do) to storing data in more flexible formats, such as documents, which NoSQL databases are designed to handle. In recent years, there has been progress in improving the ability of relational databases (like Aurora \footnote{\url{https://aws.amazon.com/rds/aurora/}, Cited on 10.09.2023 }) to scale out, which is traditionally a strength of NoSQL databases. However, in some high-scale environments, where a large amount of data needs to be managed, object stores and NoSQL databases are still often required because they offer a purer form of horizontal scaling.

\textit{Triple-stores and graph databases} provide a natural fit for storing and managing ontology-based data like RDF and OWL. While SQL database tables can encode RDF triples, and the expressive power of FOL enables specification of almost any conceptualisation, the semantics of the SQL as a means of performing graph queries is often limited as compared with a dedicated / purpose-built graph database or triple store. Additionally as mentioned above NoSQL databases greatly benefit from scaling out approaches.
Triple-stores are databases designed specifically for storing RDF triples (an example of RDF triplet shown before in Section \ref{ssec:rdf}). They typically support SPARQL, a query language for RDF, allowing for efficient querying and manipulation of the stored RDF data. Triple-stores provide the physical technological support for the practical implementation of Semantic Web applications and services (like the ones discussed in in Section \ref{ssec:semantic_web}), providing efficient storage and retrieval of RDF data.
Graph databases, on the other hand, are more general-purpose databases that use graph structures to store data. Each entity (or node) and relationship in the database can have an arbitrary number of attributes, allowing for rich and complex data models. Some graph databases support RDF and SPARQL, making them suitable for Semantic Web applications, while others use proprietary query languages. Compared to triple stores, graph databases may provide more flexibility and performance optimisations for certain types of queries and data models.

\paragraph{Database selection and integration.}
When selecting a database for storing and managing ontology-based data, it is more important to consider not only the storage of data (whether a SQL database can or cannot store JSON data), but as or more importantly, the semantics of the data (how the data is typed and queried). Almost any database can store RDF data or document (there is usually a mapping of some kind). The more important question is whether the query and type language give themselves to this mapping. If the mapping is forced, one may be able to store data, but it may be very difficult to query or to enforce constraints. For example, one may be able to store data as JSON in a SQL column, but can one impose constraints over the structure of the data in the JSON column - and can one more easily query the data using SQL language or using a graph-based query language? 
Questions one may wish to consider when adopting a database are: first, does the type system enable you to model data of your domain and enforce constraints; and second, does the query language fit the shape of your domain and are the queries easy to write and understand, once written? Quite often, it is possible to store data using a poorly fitted databases; but cracks emerge in data that is poorly constrained and queries that are hard to read or understand.




\section{Knowledge engineering: wind energy domain review} 
\label{sec:KE_in_WindEnergy}
In this section, we review and evaluate existing knowledge engineering related efforts and initiatives in the wind energy sector. This work was centred on the following four questions, which are discussed in more detail in the next sections: (1) Who are the data users and producers in the wind energy domain? (2) Which semantic artefacts relevant to these data users and producers in the wind energy domain already exist? (3) What are the gaps and overlaps in existing semantic artefacts and to what extent existing artefacts gained domain or industry adoption? (4) What types of digital twins and decision support systems have been developed so far in wind energy domain and how these systems can be improved by applying knowledge engineering methods? 

\subsection{(1) Who are the data users and producers in the wind energy domain?}
\paragraph{Scope of the domain}
This part began with establishing the scope of the wind energy domain. While we expected that the semantic artefacts we would find would focus on the fields of engineering and atmospheric science, we wanted to be aware of user communities and stakeholders outside these fields, who might use data to inform their decisions. Consequently, we were purposeful towards being inclusive of all the roles and touch-points with the domain. An understanding with taking this approach is that there would be semantic artefacts and data models from other domains that could interact with ones specific to wind energy. It was important to us that we consider the multidisciplinary nature of activities undertaken by various stakeholders and their interactions. Recognising these relationships could influence how ontologies are designed in areas where gaps or overlaps exist. Moreover, these aspects shape ontology reuse and alignment activities. 
In the absence of universally accepted classifications for the roles and activities within wind energy domain, we bounded our scope to specific stages in the life cycle of wind energy assets. In particular, we have adopted the same stages as the ones used by the authors of the \cite{Barber2023usecasesWE} in their analysis of various stakeholder "pain points" related to the digitalisation process in wind energy:
\begin{enumerate}[label=\Alph*.]
\footnotesize
    \item Wind turbine design
    \item Wind farm planning
    \item Wind farm operation
    \item Project selling / buying
    \item End of life
    \item General
\end{enumerate}
This classification is sufficiently top-level to include activities with a narrower scope such as wind resource assessment or wind turbine maintenance.   
Roles and activities related to wind turbine design and wind farm planning were selected as the initial bound to the domain of inquiry, as these provided a definite point in time in which data is present in the life cycle. To close our scope, we selected roles related to the end of life stage. When looking at other power alternatives such as hydro, nuclear, and fossil, the decommissioning stage creates new data such as the impact on the industry and environment. The same should be expected for wind energy as turbines age and build materials and designs are enhanced, climate change impacts the atmospheric conditions at existing sites, and other energy generation technologies come to market.\\
It is important to note that some of the roles and activities present in resulting search space are not entirely wind energy specific, such as environmental reviews and power consumption analysis. The search for semantic artefacts related to such roles and activities was not as extensive.

\paragraph{Type of data users}
For defining the data users and consumers in our scope of the wind energy domain, we supplemented our own domain expertise with information from the United States Bureau of Labor Statistics that described careers in wind energy \cite{blsCareersWind} and sources describing stakeholder analysis in real and theoretical wind energy scenarios \cite{bef2017stakeholderEnergy, vivero2023stakeholder}. These scenarios were of applications of onshore and offshore wind energy in Europe. A search in the Scopus database for the query (“wind energy” AND “stakeholder analysis”) provided only seven results, which focused mostly on socioeconomic effects \cite{MendietaVicua2020energypolicy, Vicuna2020social, HuescaPerez2016social}, decision making in a regulated industry \cite{Rosenberg2019}, synergy with agriculture sector \cite{Markovska2013agriculture}, marine biodiversity and aquaculture implications \cite{Wever2015marine, Aschenbrenner2019marine, Weber2022marine}. There were no results with a focus on wind turbine design, wind farm planning and operation, project selling, or end of life of wind energy assets.

The report by the United States Bureau of Labor Statistics mentioned in the previous paragraph included jobs that can be mostly attributed to the OEMs, wind power project developers, and energy producers (see Table \ref{tab:we_jobs}). A more inclusive (but not exhaustive) classification provided by a stakeholder analysis for wind energy project assessment and planning phases in a European context \cite{bef2017stakeholderEnergy} included the following:
\begin{enumerate}
\footnotesize
    \item Public authorities.
    \item Energy producers.
    \item Investors.
    \item Experts (consultants).
    \item Environmental NGOs.
    \item Professional associations.
    \item Citizen/societal groups
    \item Land owners.
    \item Wind turbine producers (OEMs).
    \item Wind power project developers.
    \item Electricity grid owners.
    \item Universities (academia).
\end{enumerate}
The majority of these stakeholders continue to interact with wind energy domain data well through the later stages of wind energy asset's life cycle, up until the end of life. 

\begin{table}[]
\footnotesize
    \centering
    \begin{tabular}{>{\hspace{0pt}}m{0.2\linewidth}>{\hspace{0pt}}m{0.25\linewidth}>{\hspace{0pt}}m{0.45\linewidth}}
         Manufacturing Phase& Engineering &  Aerospace engineers, Civil engineers,Electrical engineers, Electronics engineers, Environmental engineers, Health and safety engineers, Industrial engineers, Materials engineers, Mechanical engineers, Engineering technicians, Drafters \\
         &General Manufacturing & Machinists, Computer-controlled machine tool operators, Assemblers, Welders, Quality-control inspector, Industrial production managers\\
         &&\\
         Project Development&Management and Legal&Project Managers, Asset managers,Land acquisition specialists, Logisticians\\
         &Scientists&Atmospheric scientists, Wildlife biologists, Geologists, Environmental scientists\\
         &Construction Occupation& Construction equipment operators, Crane operators, Electricians\\
         &&\\
         Operation and Maintenance Phase&General Operation&Plant supervisors, Service technicians
    \end{tabular}
    \caption{Wind energy sector jobs per life cycle phase according to United States Bureau of Labor Statistics}
    \label{tab:we_jobs}
\end{table}

\subsection{(2) Which semantic artefacts relevant to these data users and producers in the wind energy domain already exist?}

To build a collection of semantic artefacts to review and analyse, we solicited the IEA Wind Task 43 Working Group 1 participants, a group comprised of industry, academic, and government collaborators with interests and experience in wind energy metadata\footnote{https://www.ieawindtask43.org/wg1}. A search of the SCOPUS database using the query ((taxonomy OR schema OR ontology OR "knowledge base") AND (“wind energy” OR "wind turbine" OR "wind plant"  OR  "wind power plant")) provided 202 results from scholarly literature. Chosen results from this query were selected based on having a primary focus of describing the development of an ontology or presented a clear application of an ontology in an applied setting. A search using the same query in the web search engines presented trade literature, technical reports from government agencies, and wind energy domain semantic artefacts of various expressiveness and generality. The processed results of these searches are presented hereafter. Brief summaries of wind energy domain specific artefacts are presented in Table \ref{tab:SA-WEDomain} and Table \ref{tab:we_se_purpose}. Meanwhile, cross-domain and wind energy related semantic artefacts that appeared in search queries are briefly summarised in Table \ref{tab:SA-relevant}. 

\paragraph{Review methodology.}
To perform a methodological overview of the relevant semantic artefacts, we have evaluated them with the following criteria:
(1) Context and purpose for the semantic artefact development; (2) Target audience / Role; (3) Associated activity; (4) Associated life cycle stage of the wind energy assets; (5) Semantic artefact type; (6) Alignment with other semantic artefacts (7) Technologies used. Additionally, we have assessed semantic artefacts according to semantic expressiveness, generality and granularity.
Next, we prepared a matrix that mapped roles and activities we identified within our scope of the wind energy domain with the conceptualisations that were found. Identifying the intersections between role and ontology was based on qualitatively analysing the purpose and applicability of the ontology as described by the resource or by analysis of the terms in the ontology and the list of roles. A role that did not have an identified semantic artefact would indicate a potential gap. On the other hand a roles with several distinct ontologies serving a similar purpose would indicate a lack of community adoption and dialogue. These cases required a further investigation into the possible causes of this lack in sustainable development. 
To add context to these inquiries, we have classified all semantic artefacts according to the following criteria: (1) Level of adoption; (2) Stakeholder type (Academia / Industry / Government / Standardisation Body); (3) Availability of the semantic artefact for download in some kind serialisation or as Linked Data (4) Continued development and maintenance.

\paragraph{Limitations in the analysis.}
Results from our academic literature and internet searches do not include proprietary semantic artefacts. This may constrain the conclusions of the analysis of gaps and overlaps in these conceptualisations because we are unsure of the extent that industry has identified these and developed solutions. We are also unaware of the impact of these gaps and overlaps to their data needs and operations.
An exhaustive search for semantic artefacts that included multiple energy sources in addition to wind was not performed. The assumption was that upper-level conceptualisations would not have the specificity of terms or architecture demanded by the roles in our scope of wind energy.

\paragraph{Wind energy specific semantic artefacts.}
This group of semantic artefacts obtained from SCOPUS and web search engine queries are fully wind energy domain specific as defined per scope outlined above. 
The IRPWIND initiative was intended as means to provide additional rich metadata to wind energy data sets in the context of web publishing. This was envisioned as an extension to the Dublin Core metadata model to allow for additional controlled vocabularies of terms that can be used to contextualise data sets and used for data search and retrieval purposes. For example, these terms could be used as filter tags when searching for data set in a catalogue.  
The IEA Wind Data Models were developed as part of different IEA Wind Tasks. The LiDAR ontology developed as a part of Task 32 had as a goal to facilitate analysis and exchange of data produced during measurements with various LiDAR Systems. IRPWIND and Task 32 are the only initiatives that published their semantic artefacts as Linked Data. The WindIO ontology was developed by the Task 37 group for definition of the inputs and outputs for systems engineering multidisciplinary design optimisation (MDAO) of wind turbine and plants. This resulting ontology is formalised as a YAML schema and is used to describe the structure of YAML input files for Wind-plant Integrated System Design and Engineering Model (WISDEM) software. Recently, WindIO ontology was also suggested by IEA Wind Task 55 work-group as a basis for a more general use ontology describing technical specifications and characteristics of wind turbines and power plants. This ontology will be used to define reference wind turbines and plants for the purposes of V\&V, benchmark testing, and impact assessment of novel technologies on wind plants. The WRA Data Model developed in Task 43 standardises how properties of a wind resource measurement station (e.g. latitude, longitude, anemometer serial number, installation height, logger slope, logger offset, etc.) are structured and serialised as a JSON file. This data model is described with JSON schema. 
The majority of semantic artefacts presented in academic literature are not available for download. Among these are various taxonomies of wind turbine components for reliability analysis \cite{Artigao2018taxonomyfailureanalysis}. These taxonomies are often not formalised in any modelling language. In a similar context of reliability and failure analysis, more expressive ontologies were proposed by various authors. In particular \cite{Zhou2015} for Failure Mode, Effects and Criticality Analysis (FMECA) and \cite{Papadopoulos2009} for condition monitoring. These ontologies relied on OWL for knowledge representation. 
More general, domain level ontologies and knowledge bases attempted to comprehensively capture wind energy related concepts. These authors of these knowledge bases also opted for Semantic Web technology stack, with a notable exception of WESGraph, which relied on Neo4j graph database for data storage and querying. As a consequence, the underlying top level ontology for WESGraph is not formalised with any of the commonly used ontology languages. 
Lastly, some attempts were undertaken by the industry at creating controlled vocabularies of terms used for wind turbine system SCADA and reporting data. ENTR Alliance and SCADA International created controlled vocabularies of SCADA terms in accordance to the guidelines presented in IEC 61400-25. In North America, wind turbine generation data reporting for Generating Availability Data System (GADS) follows the schema enforced by North American Electric Reliability Corporation (NERC). The conceptualisations for these three semantic artefacts are specified as lists of terms, stored along with term descriptions in a tabular form which is serialised using CSV or XLS formats.
\begin{table}
\footnotesize
\centering
\caption{Description of wind energy domain-specific semantic artefacts \label{tab:SA-WEDomain}}

\begin{tabular}{>{\hspace{0pt}}m{2em}>{\hspace{0pt}}m{0.2\linewidth}>{\hspace{0pt}}m{0.68\linewidth}}
 \hline
 & & \textbf{Brief Description} \\
 \hline
\multirow{11}{*}{\begin{sideways}IRPWIND Initiative \cite{IRPWINDmetadata}\end{sideways}}
 & ASPECT & Controlled vocabulary of variables, parameters and constants used in wind energy community. \\
 \rule{0em}{2em} & EXTRACT  & Classification of external conditions in which a wind farm operates\\
  \rule{0em}{2em} & IDEM & Classification of models used in wind energy \\
 \rule{0em}{2em} & NEAT & Taxonomical organization of research topics in wind energy which follows a typical lifecycle of wind farm development \\
  \rule{0em}{2em} & WEAR & Classification of wind turbine materials \\
  \rule{0em}{2em} & WEAVE & Classification of activities in which data are produced \\
\hline
\multirow{4}{*}{\begin{sideways}\begin{tabular}[c]{@{}l@{}}IEA Wind\\Data Models\end{tabular} \end{sideways}}
 & Task 32:\par{}  Wind LiDAR ontology & Ontological representation of wind LiDAR supporting development of modular tools and processes for wind LiDARs\\
 \rule{0em}{3em} & Task 37:\par{}  WindIO \cite{Bortolotti2022} & Wind Turbine and Wind Power Plant schemas defining input and output variables for systems engineering MDAO.\\
 & Task 43:\par{}  WRA Data Model & A JSON schema to describe and verify wind resource assement data. \\
 \hline
\multirow{25}{*}{\begin{sideways}\begin{tabular}[c]{@{}l@{}}Published as Academic Article\end{tabular}\end{sideways}} 
 &WPP Ontology \cite{Zhu2008wppontology}& Ontology built on the basis of wind power plant information model with OWL \\
 \rule{0em}{3em} & Offshore Wind \par{} Ontology (OWO) \cite{Nguyen2014} & Offshore wind domain ontology model based on the IEC 61400-25 standard. \\
 & Intelligent Fault \par{} Diagnosis of \par{} Wind Turbines \cite{Zhou2015} &  Wind Turbine fault diagnosis application ontology based on Failure Mode, Effects and Criticality Analysis (FMECA) and a knowledge base.\\
   \rule{0em}{3em}&WT Components\par{} Taxonomy \cite{Artigao2018taxonomyfailureanalysis} &A unified taxonomy of wind turbine components (including Reliawind and \cite{Reder2016}) for the purpose of reliability analysis.\\
  \rule{0em}{3em}& Condition Monitoring\par{} of Wind Turbines \cite{Papadopoulos2009} & Wind Turbine condition monitoring application ontology and a knowledge base. \\
  \rule{0em}{3em}&WT Operational\par{} States \cite{Bunte2018hybridAIOpertationalStateIdentification}&An ontology and a knowledge base of wind turbine operational states\\
  \rule{0em}{3em}&WPP Spatial\par{} Database\cite{Lungu2012wpp-sql-schema}&Conceptual schema for a SQL Database \\
  \rule{0em}{3em}&WPP Expert\par{} System \cite{Duer2017wpp-kb}& Expert system for wind power plant's equipment diagnosis\\
  \rule{0em}{3em}&Onshore WT \par{} Maintenance \cite{Strack2021}& Wind turbine maintenance task ontology for onshore wind turbines.\\ 
 \rule{0em}{2em} & OntoWind \cite{Kucuk2018} & Wind Energy domain ontology and a knowledge base. \\
 & WESgraph \cite{Quaeghebeur2020}  & Top-Level ontology and a knowledge base for the wind farm domain, implemented as a graph database. \\
 \hline
\multirow{14}{*}{\begin{sideways}\begin{tabular}[c]{@{}l@{}}Industry Developed\end{tabular}\end{sideways}} 
 &Reliawind Taxonomy& Taxonomy of wind turbine components for the purpose of reliability and FMECA analysis\\
 \rule{0em}{3em} &Power Curve\par{} Schema \cite{Clark2023PowerCurves} & A JSON schema to describe and verify WT power curve data serialised as JSON.\\
 \rule{0em}{3em} &ENTR Alliance \, OpenOA & Controlled vocabulary for SCADA data and a schema based on IEC61400-25 standard describing renewable energy variables.\\
 \rule{0em}{4em} &Global Wind Data\,  Tag-List & A comprehensive controlled vocabulary of data tags used in wind energy SCADA systems, based on the IEC 61400-25 standard. Maintianed by SCADA International and published as a spreadsheet.\\ 
 &GADS WT \par{} Reporting & CSV file schema for Generating Availability Data System (GADS) wind turbine generation data reporting used by North American Electric Reliability Corporation\\
 \hline
\end{tabular}
\end{table}
\begin{table}[]
\footnotesize
    \centering
    \begin{tabular}{>{\hspace{0pt}}m{0.25\linewidth}cc}
         \hline
         \textbf{Name}& \textbf{Life Cycle stage} & \textbf{Context /Purpose} \\
         \hline
         IRPWIND&General&\begin{tabular}{c}
                        Data management:\\
                        FAIR practices for data;\\
                        Publishing data sets on the Web;\\
                        Data search and retrieval;
                        \end{tabular} \\
         IEA Task 32:\par{}  Wind LiDAR \par{} Ontology&Wind farm planning&\begin{tabular}{c}
                                                                        Siting:\\
                                                                        LiDAR Data processing;\\
                        \end{tabular}\vspace{0.5 em}\\
         IEA Task 37:\par{}  WindIO \cite{Bortolotti2022} &Wind turbine design& \begin{tabular}{c}
                                                                                Economics:\\
                                                                                LCOE optimisation;\\
                                                                                WISDEM inputs definition;
                                                                                \end{tabular}\vspace{0.5 em}\\
         IEA Task 43:\par{}  WRA Data Model &Wind farm planning&\begin{tabular}{c}
                                                                        Siting:\\
                                                                        Wind resource assessment (WRA);\\
                                                                        WRA data sharing, processing,\\
                                                                        and interoperability;
                        \end{tabular}\vspace{0.5 em}\\
         WPP Ontology \cite{Zhu2008wppontology} & Wing farm operation&\begin{tabular}{c}
                                                                        Operations and Maintenance:\\
                                                                        WPP information systems development;
                        \end{tabular}\vspace{0.5 em}\\
         Offshore Wind \par{} Ontology (OWO) \cite{Nguyen2014} &Wing farm operation&\begin{tabular}{c}
                                                                                    Operations and Maintenance:\\
                                                                                    WPP Management;\\
                                                                                    Data integration;\\
                                                                                    Knowledge sharing;\\
                                                                                    \end{tabular}\vspace{0.5 em}\\
         Diagnosis of \par{} Wind Turbines \cite{Zhou2015} &Wind farm operation&\begin{tabular}{c}
                                                                                    Operations and Maintenance:\\
                                                                                    FMECA Analysis; DSS;
                                                                                    \end{tabular}\vspace{0.5 em}\\
         WT Components\par{} Taxonomy \cite{Artigao2018taxonomyfailureanalysis}, \par{}  Reliawind Taxonomy  &Wind farm operation&\begin{tabular}{c}
                                                                                    Operations and Maintenance:\\ Reliability Analysis;\\
                                                                                    Failure localisation on component basis;\\
                                                                                    \end{tabular}\vspace{0.5 em}\\
         Condition Monitoring\par{} of Wind Turbines \cite{Papadopoulos2009} &Wind farm operation&\begin{tabular}{c}
                                                                        Operations and Maintenance:\\
                                                                        Condition Monitoring; DSS;\\
                                                                        \end{tabular}\vspace{0.5 em}\\
         WT Operational\par{} States \cite{Bunte2018hybridAIOpertationalStateIdentification} &Wind farm operation&\begin{tabular}{c}
                                                                                    Operations and Maintenance:\\
                                                                                    WPP operation and control;
                                                                                \end{tabular}\vspace{0.5 em}\\
         WPP Spatial\par{} Database\cite{Lungu2012wpp-sql-schema} &Wind farm planning&\begin{tabular}{c}
                                                                        Siting:\\
                                                                        WRA; Spatial planning;\\
                                                                        Querying GIS Database; DSS;
                                                                        \end{tabular}\vspace{0.5 em}\\
         WPP Expert\par{} System \cite{Duer2017wpp-kb} &Wind farm operation&\begin{tabular}{c}
                                                                            Operations and Maintenance:\\
                                                                            WPP operation and control;\\
                                                                            Expert system development;
                                                                            \end{tabular}\vspace{0.5 em}\\
         Onshore WT \par{} Maintenance \cite{Strack2021} &Wind farm operation&\begin{tabular}{c}
                                                                            Operations and Maintenance:\\
                                                                            Condition-oriented maintenance;\\
                                                                            Maintenance reports digitisation\\
                                                                            and SCADA information integration;
                                                                            \end{tabular}\vspace{0.5 em}\\
         OntoWind \cite{Kucuk2018} &General&\begin{tabular}{c}Knowledge management:\\
                                                              Wind Energy KB Development
                                            \end{tabular}\vspace{0.5 em}\\
         WESgraph \cite{Quaeghebeur2020} &General&\begin{tabular}{c}Knowledge management:\\
                                                              Wind Energy KB Development
                                            \end{tabular}\vspace{0.5 em}\\
        Power Curve\par{} Schema \cite{Clark2023PowerCurves} &General&\begin{tabular}{c}Data management:\\
                                                              Automating workflows;\\
                                                              Application development;
                                            \end{tabular}\vspace{0.5 em}\\                               
         ENTR Alliance\par{} OpenOA &Wind farm operation&\begin{tabular}{c} Operations and Maintenance:\\ 
                                                                WT Supervision and Control;\\
                                                                Condition monitoring;\\
                                                                Operational analysis;\\
                                                                SCADA information management;
                                                                \end{tabular}\vspace{0.5 em}\\
         Global Wind Data\par{}  Tag-List &Wind farm operation&\begin{tabular}{c}
                                                                Operations and Maintenance:\\ 
                                                                SCADA information management;
                                                                \end{tabular}\vspace{0.5 em}\\
         GADS WT \par{} Reporting  &Wind farm operation&\begin{tabular}{c}Operations and Maintenance:\\ 
                                                                Energy generation availability data reporting;
                                                                \end{tabular}\\
         \hline
    \end{tabular}
    \caption{Wind energy domain semantic artefacts and stakeholder use cases}
    \label{tab:we_se_purpose}
\end{table}
%
%
%
\begin{table}
\footnotesize
\centering
\caption{Description of cross-domain and wind energy overlapping domains semantic artefacts \label{tab:SA-relevant}}

\begin{tabular}{>{\hspace{0pt}}m{2em}>{\hspace{0pt}}m{50pt}>{\hspace{0pt}}m{360pt}}
 \hline
 & & \textbf{Brief Description} \\
 \hline
 \multirow{12}{*}{\begin{sideways}\begin{tabular}[c]{@{}l@{}}Upper and Mid Level\end{tabular}\end{sideways}} 
& BFO  & Basic Formal Ontology (BFO) is a top-level ontology that provides a foundational framework for organizing and structuring domain-specific ontologies. It aims to promote interoperability and integration across different domain ontologies by providing a common set of basic categories and relationships.\vspace{0.5em}\\
&CCO & Common Core Ontologies (CCO) is a collection of 12 ontologies that comprise mid-level extension of BFO. The CCO provide semantics for concepts and relations that are used in most domains of interest, such as concepts from Units of Measure Ontology, Event Ontology, Time Ontology. Many domain ontologies are aligned with CCO. Among them are Aircraft Ontology, Occupation Ontology, Sensor Ontology, etc.\vspace{0.5em}\\
& Dublin Core & Dublin Core is a simple, flexible, and extensible metadata standard for describing a wide range of resources, including digital and physical assets. It consists of a set of 15 core elements (e.g., Title, Creator, Subject, etc.) that can be used to describe resources in a consistent and structured manner, facilitating resource discovery and interoperability.\vspace{0.5em}\\
& Schema.org & A shared vocabulary for structured data markup on web pages to improve search engine results and discoverability.\\

 \hline
 \multirow{3}{*}{\begin{sideways}\begin{tabular}[c]{@{}l@{}}W3C\end{tabular}\end{sideways}} 
& SKOS & A W3C data model recommendation for expressing controlled vocabularies, taxonomies, and thesauri.\vspace{0.5em}\\

& PROV-O & The PROV Ontology (PROV-O) is an ontology that provides a vocabulary for expressing provenance information.\vspace{0.5em}\\

& SSN-XG & Semantic Sensor Network (SSN-XG) ontology describes sensors, actuators and observations, and related concepts. Domain concepts, such as time, locations, are intended to be included from other ontologies via OWL imports. \vspace{0.5em}\\

\multirow{4}{*}{\begin{sideways}\begin{tabular}[c]{@{}l@{}}W3C Example\\Ontologies\end{tabular}\end{sideways}} 
& AWS & Agriculture Meteorology example, Ontology for Meteorological Sensors showcasing the ontology developed by the W3C Semantic Sensor Networks incubator group (SSN-XG). \vspace{0.5em}\\

& CF & Climate and Forecast (CF) metadata conventions vocabulary encoded with OWL \vspace{0.5em}\\

& WEATHER & Linked Sensor Data - Weather station is an example ontology for weather data publication on the LOD using SOSA/SSN ontology \vspace{0.5em}\\

& WM30 & An example of the SSN-XG sensor ontology used to describe a specific sensing device, the Vaisala WM30, which measures wind speed and wind direction.\\
\hline
 \multirow{8}{*}{\begin{sideways}\begin{tabular}[c]{@{}l@{}}Observations and \\Measurements\end{tabular}\end{sideways}} 
& SciDATA & SciData is a data model for scientific data that provides an ontologically defined framework for organizing and linking (with JSON-LD) both the data and metadata from scientific experiments, calculations, and theories.\vspace{0.5em}\\
& OBOE & The Extensible Observation Ontology (OBOE) is a formal ontology for capturing the semantics of scientific observation and measurement. The ontology supports researchers in adding detailed semantic annotations to scientific data, thereby clarifying the inherent meaning of scientific observations. \vspace{0.5em}\\
& I-ADOPT & An ontology designed to facilitate interoperability between existing variable description models (including ontologies, taxonomy, and structured controlled vocabularies). \vspace{0.5em}\\
& OM & Ontology of units of Measure (OM) is an ontology focused on units, quantities, measurements, and dimensions relevant to scientific research\vspace{0.5em}\\
\hline
\multirow{8}{*}{\begin{sideways}\begin{tabular}[c]{@{}l@{}}Domain Level\end{tabular}\end{sideways}} 
& SWEET & Semantic Web for Earth and Environmental Terminology (SWEET) is a foundational ontology that contains over 6000 concepts organized in 200 ontologies expressed in OWL.  SWEET is a highly modular, general-purpose ontology suite designed to represent Earth and environmental science concepts and their relationships.\vspace{0.5em}\\
& CF Metadata \- Conventions & Climate and Forecast (CF) standardized set of metadata elements for describing climate and forecast data stored in netCDF files. The conventions aim to facilitate data sharing, discovery, and interoperability in the climate and forecasting communities. \vspace{0.5em}\\ 
&SLAKS&Semantic LAminated Composites Knowledge management System (SLACKS) \cite{Premkumar2014} based on suite of ontologies for laminated composites materials and design for manufacturing (DFM)\vspace{0.5em}\\
 \hline
 \multirow{8}{*}{\begin{sideways}\begin{tabular}[c]{@{}l@{}}Energy Domain\end{tabular}\end{sideways}} 
 & OEO & Open Energy Ontology (OEO) \cite{Booshehri2021}  is an energy system modelling domain ontology. \vspace{0.5em}\\
 
 & EKG & Energy Knowledge Graph (EKG) \cite{Chun2018} is an upper level ontology for the integration of knowledge resources in energy systems\vspace{0.5em}\\

 &EDF PPO& Electricity of France (EDF) Power Plant Ontology and knowledge base \cite{DourgnonHanoune2010}.\vspace{0.5em}\\
 
 & GCIEO & Global City Indicator Energy Ontologies (GCIEO) \cite{Komisar.2008} is a standard ontology for Semantic Web based representations of general knowledge for the Energy Theme indicators (ISO 37120 ). \vspace{0.5em}\\

 \hline
\end{tabular}
\end{table}

\paragraph{Cross-domain and wind energy activities related domains.}
This group of semantic artefacts, which is not entirely contained within the scope of domain of interest, appeared among the results of the SCOPUS and web search engine queries due to their cross-domain nature and applications in wind energy (and as result would match to keywords like "wind turbine" or "wind energy"). These semantic artefacts can be attributed to the domains that overlap with wind energy such as environment and meteorology, sensing, structural health monitoring, material sciences, energy etc. As mentioned before, no targeted search and review was performed for each of the overlapping domains, hence the presented list is not exhaustive. For example, Semantic LAminated Composites Knowledge management System (SLACKS) was developed specifically for the wind turbine blades design use case. However, besides this specific case, a multitude of material ontologies and knowledge bases exist.
Table \ref{tab:SA-relevant} presents a summary of the semantic artefacts reviewed for this work. Most these have been adopted by various communities and are instrumental for inter-disciplinary collaborations.
The table does not include semantic artefacts that have not seen the wide-spread adoption, either due to their "in-development" status or when superseded by more recent efforts. For example, a Structural Health Monitoring (SHM) Ontology \cite{Tsialiamanis2021} was recently proposed to facilitate knowledge sharing, application, and reusability for SHM projects. However, it has not been yet validated and published. At the same time some renewable energy domain ontologies such as OpenWatt\cite{Lamanna2014} are no longer supported as the knowledge has been subsumed by knowledge bases such as Open Energy Ontology (OEO)\cite{Booshehri2021}.

 For the sake of completeness, several upper and mid level ontologies have been reviewed, such Basic Formal Ontology (BFO) and Common Core Ontologies (CCO) as many domain specific semantic artefacts developed by communities outside of wind energy tend to align with some upper level ontology.  It should be noted that there are many upper level ontologies that have been developed by various authors with different focus. For example, Dublin Core was developed in context of metadata standards and description of web resources, such as publications, data sets, images etc. Another example is Schema.org: a commonly adopted ontology for describing resources on the web, initially developed for e-commerce scope by a consortium of Google, Microsoft, Yahoo and Yandex. A comprehensive overview and evaluation of upper level ontologies was performed by \cite{Partridge2020TopLevelOntologies}.
 Ontologies and data model recommendations developed by W3C such as Simple Knowledge Organisation System (SKOS), PROV-O and Semantic Sensor Network (SSN-XG) have gained widespread adoption due to pioneering work of W3C on "web of data" and its role in Semantic Web technology stack development. SKOS provides a system for creation of taxonomies, controlled vocabularies and thesauri creation within this technology stack, following the principles of Linked Data. In wind energy context, IRPWIND initiative used SKOS in creation of their taxonomies. The PROV Ontology (PROV-O) is an ontology that provides a vocabulary for expressing provenance information, which can be important in wind energy context for data governance purposes given the multiplicity of stakeholders and complexity of the systems producing the data. Several examples of Semantic Sensor Network (SSN-XG) ontology have been developed specifically for meteorological sensors, which clearly overlaps with types of data generated in the context wind energy activities. More generally, data generated during observation and measurement activities can be formalised with data models like SciDATA and described with ontologies like Extensible Observation Ontology (OBOE), Ontolgy of Measurement Units (OM) and I-ADOPT. In addition, semantic artefacts such Semantic Web for Earth and Environmental Terminology (SWEET) and Climate and Forecast (CF) Metadata conventions from weather and environmental domains share significant terminological overlap for describing observations and measurements related to siting activities.
 Several energy domain ontologies and knowledge bases like Open Energy Ontology (OEO), Energy Knowledge Graph (EKG), Global City Indicator Energy Ontologies (GCIEO), and Electricity of France (EDF) Power Plant Ontology include some wind energy concepts. As a result, these semantic artefacts appeared among the searches performed for this review. 
 
\subsection{(3)  What are the gaps and overlaps in existing semantic artefacts and to what extent existing artefacts gained domain or industry adoption?}
The results of the search queries and their subsequent analysis has revealed the following:
\begin{itemize}
    \item The majority of existing semantic artefacts pertain to the wind farm operation life cycle stage, and especially to the activities related to failure and reliability analysis of wind turbines.
    \item There appears to be no semantic artefacts developed specifically with the context of project selling / buying or end of life stages. 
    \item Existing semantic artefacts have not gained high adoption by domain experts and there is no common domain-level ontology that is accepted by the community.
    \item There is no alignment to upper level ontologies or between semantic artefacts within the wind energy domain. Similarly, there no alignment or re-use of semantic artefacts from domains overlapping with wind energy in their data generation and producing activities. 
    \item There is significant corpus of taxonomies and vocabularies that has not been formalised with any modeling language. Many activities still rely on manual data processing. 
\end{itemize}

\begin{figure}[t]
    \centering
    \includegraphics[width=\textwidth , trim=60px 240px 60px 60px, clip]{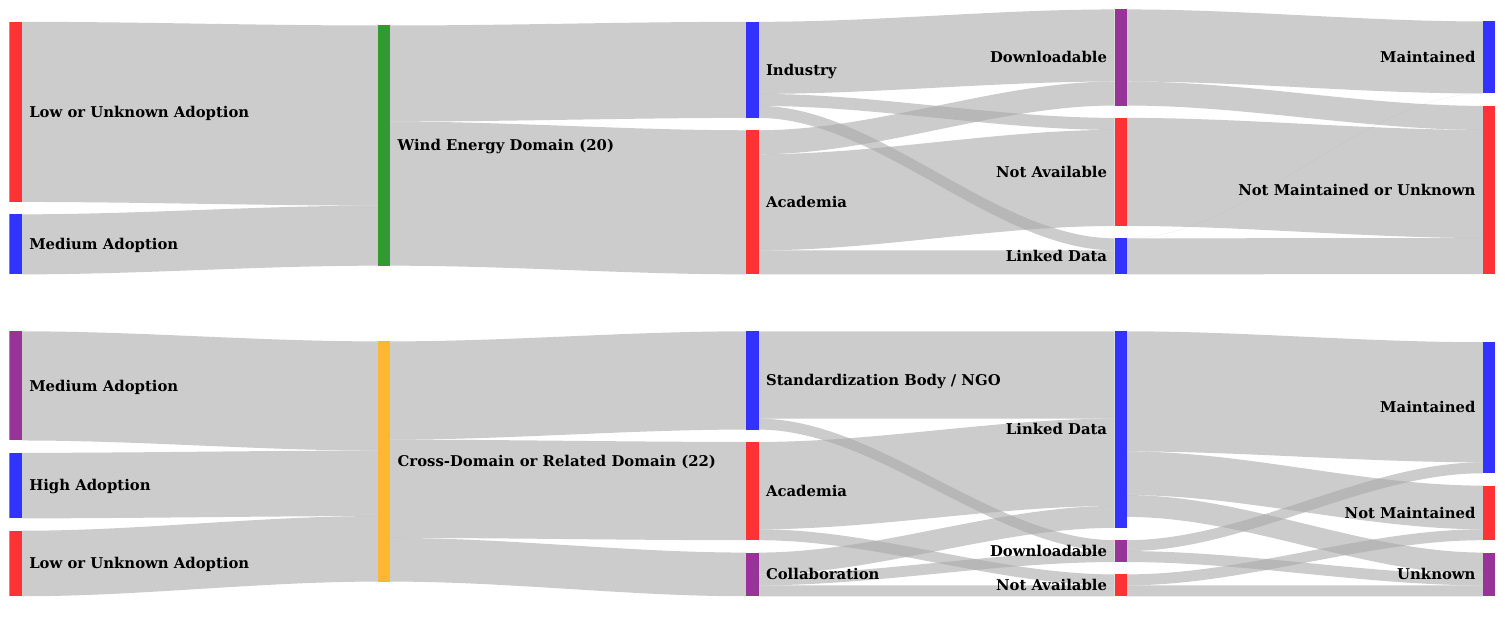}
    \caption{Analysis of semantic artefacts adoption levels. Low adoption levels in wind energy domain can be attributed to low availability and lack of active development.}
    \label{fig:sankey-we}
\end{figure}

These conclusions can be illustrated with the example of taxonomies developed for the purpose of failure analysis. The authors of \cite{Artigao2018taxonomyfailureanalysis} manually unified and aligned 13 different wind turbine component taxonomies (none of which was made available using some kind of standard formalisation). Following this trend, one of the more recent wind turbine failure analysis performed by \cite{SanchezFernandez2023} once agian manually mapped failure and maintenance records to a new WT taxonomy based on the Standard Reference Designation System for Power Plant (SRD-PP). This lack of alignment and reuse is also highlighted in \cite{Leahy2019dataqualityissues} as the authors remark that the absence of unified standards for turbine taxonomies, alarm codes, SCADA operational data and maintenance and fault reporting significantly hinders the wind turbine condition monitoring and reliability analyses. Such a situation can be partially attributed to the fact that the existing conceptual models in wind energy and related domains are not maintained and are not published following the LD  principles, as can be observed from the Sankey diagram in Figure \ref{fig:sankey-we}.\\ 
Out of 19 reviewed existing wind energy domain semantic artefacts, six were downloadable in some kind of serialisation, and three were not even available. Such a situation results in low adoption and lack of further development in a negative feedback cycle. This issue is not unique solely to the wind energy domain, but also for many technological sciences. Meanwhile, this is generally not the case for cross-domain and top-level semantic artefacts. Such artefacts are widely used by few communities spearheading open science principles adoption such as BioMed community. Hence, there is a strong need for a holistic approach: a framework for community development and maintenance of semantic artefacts, in addition to a platform for semantic artefact hosting and usage, which is discussed more in Section \ref{sec:discussion}.

\subsection{(4) What types of digital twins and decision support systems have been developed so far in the wind energy domain, and how can these systems be improved by more widespread adoption of common semantic artefacts?}
\begin{figure}[b]
    \centering
    \includegraphics[width=0.9\textwidth , trim=120px 300px 120px 0px clip]{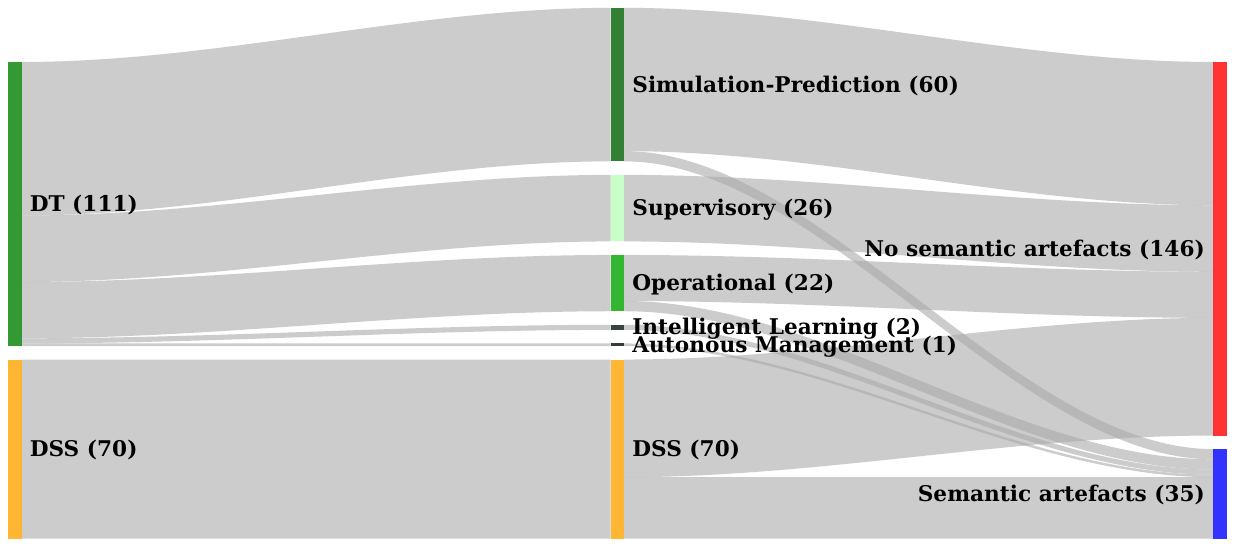}
    \caption{Semantic artefacts adoption in digital twins and decision support systems based on literature review}
    \label{fig:sankey-dt}
\end{figure}

For the purpose of this section, a systematic review of the publications regarding decision support systems and digital twins in the wind energy domain has been performed. A search of the Scopus database using the query (("decision support system" OR "expert system" OR "digital twin" ) AND (“wind energy” OR "wind turbine" OR "wind plant"  OR  "wind power plant")) yielded  532 results on first of August 2023. After removing "false positives", i.e. papers that did not actually relate to any of the queried topics, and selecting the results relevant to the question posed, the remaining  181 results have been compiled in Figure~\ref{fig:sankey-dt} and classified based on the modelled component or assembly, as well as the functional level of the DT system (Supervisory, Operational, Simulation-Prediction, Intelligent-Learning,  Autonomous-Management) \cite{Wagg2020}. The levels differ in the integration of datasets, starting from signal conditioning, including metadata, up to using ontologies. Further distinctions are made by e.g. the level of integration of numerical models.

Out of the 181 results, 111 of them cover topics related to digital twin implementations, and the remaining are related to decision support systems.
Most digital twin implementations were found to belong to the functional levels "Supervisory" (26 out of 111), "Operational" (22 out of 111) or "Simulation-Prediction" (60 out of 111). Only three papers belong to the functional levels "Intelligent-Learning"~\cite{Chatterjee2020ILDT, Li2021ILDT} and "Autonomous-Management"~\cite{ChaveroNavarrete2019AMDT}.


Generally, it can be seen that there is a lack of adoption of semantic artefacts in the research of digital twin and decision support systems, reflected by the low number of papers that use them (35 out of 181). The research of digital twins is focused on the details of the analytic
methodologies that are implemented in the digital twin rather than focusing on utilising semantic artefacts.
Therefore it can be concluded that there is a low level of adoption of semantic artefacts and of digital twins with intelligent or autonomous features in the wind energy sector.
\section{Discussion}
\label{sec:discussion}
As digitalisation within the wind energy sector matures, we anticipate the current corpus of ontologies, schemas, and data models to develop and grow. The wind energy knowledge engineering ecosystem, including semantic artefacts, tools and applications, and actors will evolve to enable and support comprehensive data management and analysis throughout the wind energy sector. 
There are, however, several essential requisites for this ecosystem to be healthy and thriving, which could be identified in this work and which are described below. They are divided into three categories: (1) Organisation and Diversity, (2) Productivity and (3) Resilience.

\paragraph{Organisation and Diversity.}
The digitalisation process should cater to the distinct needs of the manifold stakeholders in the wind energy landscape. 

\begin{itemize}
    \item \textbf{Stakeholder analysis:}
    A thorough taxonomy of wind energy stakeholders and their activities will aid in discerning use-cases. This builds on the groundwork laid by \cite{Barber2023usecasesWE}, but broadens the scope to incorporate entities like public groups, NGOs, and governments. Currently, based on our review, the digitalisation process seems to overlook interests and use cases of the most of stakeholder groups with the exception of wind turbine producers (OEMs) and energy producers.
    \item \textbf{Inclusive stakeholder engagement:} 
    Creation of public forums and collaborations are vital for the community growth and development. Examples of such activities are the WeDoWind framework, which incentivises data sharing via "challenges" set by data providers \cite{Barber.2022, Barber.2023a, Barber.2023b}, and IEA Wind Task 43, which aims to accelerate digital transformation in the wind sector by acting as a catalyst of open collaboration\footnote{https://www.ieawindtask43.org/}. 
    \item \textbf{Comprehensive digitalisation of the entire wind energy sector:} 
    It is pivotal to consider every stage of the wind energy project life cycle. Currently, emphasis largely revolves around the Operation and Maintenance phase. Task and Application ontologies catering to other stages, including end-of-life and financial activities, must be developed.
    \item \textbf{Balancing expressiveness with simplicity:}
    Depending on the use case, the semantic artefacts might require different expressiveness or different paradigm (e.g. OWA vs. CWA) adoption. Not all activities require a definition of a fully developed ontology. In many cases, a controlled vocabulary of terms, a taxonomy or a schema would be more adequate. This should be assessed by the community during the initial stages of the semantic artefact development.
\end{itemize}

\paragraph{Productivity.}
Productivity encapsulates the community's prowess in generating and innovating new semantic artefacts and tools that can respond to ever-evolving sector needs. The ability to continually produce these new resources ensures that the sector not only remains at the forefront of technology but also proactively addresses emergent challenges, adding significant value to all stakeholders. In addition to creation of new semantic artefacts and AI-tools, a productive ecosystem should streamline existing workflows and maintain effectiveness and efficiency of data and knowledge management processes.

\begin{itemize}
    \item \textbf{Generation of new semantic artefacts and standards:} Development of new semantic artefacts of various expressiveness based on stakeholder use cases (or "pain points") is paramount for successful digitalisation.
    This is aligned with IEA Wind Task 43 digitalisation activities, within which the creation of ontology and schema development groups is envisioned. For reference, BioPortal hosts 1065 semantic artefacts, which is almost two orders of magnitude more than number of wind energy domain semantic artefacts reviewed in this work.
    Some of the ontologies and schemas already proposed for development within IEA Wind Task 43 are reported in Table \ref{tab:ontology_development}
    \item \textbf{Cross-pollination:}
    The increased productivity can be achieved by utilising expertise from diverse groups within the wind energy domain, as well as other industries that are successfully undergoing digitalisation, such as bio-med and e-commerce through ontology re-use and alignment. As highlighted by this review, currently there are no efforts that seek to re-use and align semantic artefact within the wind energy domain. 
    Wind energy community should consider top level, cross-domain, and related domain ontologies that are already well established and accepted within respective communities for ontology reuse.   
    The infra-domain alignment of wind energy semantic artefacts can significantly improve the efficiency and quality of data analysis. For instance, as has been noted in this present work, the alignment of various taxonomies of wind turbine parts can be performed with relatively minimal effort, while offering immediate payback.
    The alignment of wind energy domain semantic artefacts with relevant ontologies outside of the wind energy domain also may offer significant benefits. For example, alignment with Global City Indicator Energy Ontologies can be useful for use cases relative to public and government types of stakeholders.
    \item \textbf{Information access and transparent decision-making}: Creation of web pages and applications that interact with wind energy domain knowledge bases and provide various stakeholders with information of interest. Here the topics such as ontology based data integration and data management play an important role. 
    \item \textbf{Automation:} Development of new tools for AI systems. This, as well, is in line with IEA Wind task 43 roadmap.  Workflow development.     
\end{itemize}

\begin{table}[]
\footnotesize
    \centering
    \begin{tabular}{>{\hspace{0pt}}m{0.24\linewidth} >{\hspace{0pt}}m{0.14\linewidth} >{\hspace{0pt}}m{0.46\linewidth}}
        \hline
        \textbf{Name}& \textbf{Expressiveness} & \textbf{Description and purpose} \\
        \hline
         Vocabulary of SCADA Terms &Vocabulary& Controlled vocabulary of terms created with Semantic Web technologies and published online. Assigns URIs to the SCADA terms in accordance with 61400-25. These URIs can be used to describe SCADA data with RDF.\\
         &&\\
         Wind Turbine System \par Components & Taxonomy&A classification of wind turbine components using SKOS data model. This classification can be adopted for various tasks and applications (e.g. reliability analysis) requiring minimal ontological commitment.\\
         &&\\
         Wind Energy Project \par{} Life Cycle Stages&Taxonomy&A classification of a wind energy project life cycle stages. This classification can be use in any context that requires an organisation of data based on life cycle stage criteria.\\
         &&\\
         Wind Energy Stakeholders& Taxonomy&A classification of wind energy domain actors and roles. Forms a basis for a comprehensive analysis of stakeholder needs. Additionally, can be used for Dublin Core Audience\tablefootnote{http://purl.org/dc/terms/audience} types.\\
         &&\\
         Wind Energy Activities& Taxonomy&Improved and more comprehensive taxonomy of wind energy activities based on WEAVE. Can be used for use-case driven analysis of stakeholder needs and in definition of Dublin Core Subject\tablefootnote{http://purl.org/dc/elements/1.1/subject} property.\\
         &&\\
         Blade Damage &  Taxonomy& A classification and controlled vocabulary to describe types of wind turbine blade damage. Possible use cases include, but not limited to, uniform monitoring and maintenance reporting, data tagging (e.g. photographic data, SHM data etc), context aware data analysis etc.\\
         &&\\
         Airfoil Data Model& Schema&A schema specifying wind turbine airfoil characteristics. The primary purpose is serialisation, validation and exchange of data. This can be used in the development of software applications that model airfoil aerodynamics and workflows that involve aerodynamic data.\\
         &&\\
         Wind Turbine System \par Specification & Schema&A schema specifying wind turbine characteristics used for serialisation, validation, exchange of data, and software application development.\\
         &&\\
         Wind Turbine System \par Sensors&Schema&Description of sensors that are installed on wind turbines for monitoring purposes. This metadata can be included alongside the data generated by the sensor in machine readable format.\\
         &&\\
         Power Curve Data & Schema &Specification of wind turbine power curve data with JSON Schema. The primary use of the schema is during serialisation, validation and exchange of data.\\
         &&\\
         Wind Turbine Coordinate \par Systems &Schema /\par{} Ontology & Formal description and URIs for coordinate systems defined for a generic wind turbine in accordance with IEC/TS 61400-13. These coordinate systems can be used to define the location of the wind turbine components, damages, or installed sensors.\\
         &&\\
         Wind Turbine System \par Ontology (WTSO)& Ontology& Ontological representation of a wind turbine. Creation of Knowledge Bases containing information about wind turbine systems. Lower level conceptualisations can be aligned with this ontology.\\
         
    \end{tabular}
    \caption{Semantic artefacts proposed for development within IEA Wind Task 43}
    \label{tab:ontology_development}
\end{table}

\paragraph{Resilience.}
Resilience ensures that the ecosystem will adapt and evolve in the face of challenges, ensuring its longevity and relevance. For the wind energy sector, it involves creating robust and flexible knowledge infrastructures that can accommodate technological advancements, shifting stakeholder needs, and external disruptions.

\begin{itemize}
    \item \textbf{FAIR principles:} Adherence to FAIR principles can be facilitated by use of Semantic Web technology stacks and Linked data. An example of such an effort, which would directly benefit the wind energy community, is the creation of an ontology hosting catalogue for the technology sciences. 
    \item \textbf{Maintenance:} Adoption of Free Open Source Software (FOSS) community practices and technologies (such as Git) can aid significantly in ensuring long-term support and the sustainable development of knowledge engineering applications for the wind energy sector. This is vital, as many of the existing semantic artefacts reviewed are still under development, while others would benefit from further development and improvement after methodological evaluation and assessment.  
\end{itemize}

Aspects related to culture and coopetition are also important to consider for a healthy and thriving wind energy knowledge engineering ecosystem. These aspects relate to some of the key challenges in the digitalisation of wind energy recently introduced in a review paper \cite{Clifton2022}. The topic of culture involves, for example, developing and maintaining collaborative organisational cultures, combining staff skills in new ways, enhancing communication skills, developing change processes, and increasing diversity. The topic of coopetition involves enabling cooperation, collaboration, and competition between organisations. This means working together to create marketplaces or business opportunities that would not otherwise exist and that are mutually beneficial. A discourse on these elements, however, transcends the ambit of this knowledge engineering review.

\section{Conclusions}
\label{sec:conclusions}
The wind energy sector is amidst the global transformative phase of increased automation and rapid digitalisation.
While the digital transformation is paving the way for advancements like AI-powered digital twins and decision support systems, in wind energy domain, challenges remain, particularly in the realm of converting raw data into meaningful domain knowledge that is both humanly and machine understandable. A significant part of this challenge is the lack of widespread expertise and tools in data management and knowledge engineering, leading to underutilised, undervalued, and fragmented data often void of context. 
The current work has attempted to bridge this knowledge gap by shedding light on the relevance and utility of knowledge engineering for the wind energy domain. It has presented a coherent synthesis of existing works in knowledge engineering and representation, tailored for wind energy experts. 
Through a systematic review, this study also underscores the pressing need for an inclusive approach that caters to a wide range of stakeholders, an increased generation of semantic artefacts and data management tools, and a robust infrastructure with a focus on sustainable development to ensure resilience.
However, true progression can only be realised when collaborative efforts within the wind energy community are intensified. This involves not just internal coordination but also leveraging insights from other sectors that have already navigated their digital transformation and have effectively utilised knowledge engineering methods and technologies. Existing efforts like IEA Wind Task 43 are commendable initiatives in this direction providing a foundational starting point. Embracing these initiatives and fostering collaboration will undoubtedly steer the wind energy sector towards a future that maximizes the potential provided by digital transformation.



\vspace{6pt} 



\section*{CRediT statement}
Conceptualization, Y.M. and J.D.; methodology, Y.M. and J.D.; investigation, Y.M., T.C., J.D., M.W., and J.Q.; data curation, Y.M.; writing---original draft preparation, Y.M.; writing---review and editing, Y.M., T.C., J.D., M.W., C.H., J.Q., A.M.S., I.A., E.C., and S.B.; visualization, Y.M.; supervision, E.C. and S.B.; project administration, S.B.; funding acquisition, S.B. All authors have read and agreed to the published version of the manuscript.

\section*{Funding}
A portion of this work is funded by the BRIDGE Discovery Programme of the Swiss National Science Foundation and Innosuisse (Project Number 40B2-0\_187087).\\
A portion of this work was supported by the Wind Data Hub funded by U.S. Department of Energy, Office of Energy Efficiency and Renewable Energy's Wind Energy Technologies Office operated and maintained by Pacific Northwest National Laboratory at \url{https://a2e.energy.gov.}


\section*{Conflicts of interest}The authors declare no conflict of interest. The funders had no role in the design of the study; in the collection, analyses, or interpretation of data; in the writing of the manuscript, or in the decision to publish the~results.


\section*{Abbreviations}
The following abbreviations are used in this manuscript:\\

\noindent 
\begin{tabular}{@{}ll}
FOL & First Order Logic\\
DL & Description Logic\\
OWA & Open World Assumption\\
CWA & Closed World Assumption\\
SQL & Structured Query Language\\
SPARQL & SPARQL Protocol and RDF Query Language \\
W3C & World Wide Web Consortium\\
URI & Uniform Resource Identifier\\
LD & Linked Data\\
RDF & Resource Description Framework\\
RDFS & Resource Description Framework Schema\\
OWL & Web Ontology Language\\
SKOS & Simple Knowledge Organisation System\\
DT & Digital Twin\\
DSS & Decision Support System\\
KBS & Knowledge Based Systems\\
ML & Machine Learning\\
AI & Artificial Intelligence\\
SCADA & Supervisory Control and Data Acquisition\\
WT & Wind Turbine\\
WPP & Wind Power Plant\\
OEM & Original Equipment Manufacturer
\end{tabular}

\newpage
\appendix
\section{Ontology Languages}
\label{app:ontology_languages}
Here we provide several examples of statements made in modelling languages, that are commonly used in knowledge engineering context. 
\paragraph{RDF Schema (RDFS)} 
RDF Schema provides a basic type system for RDF. It introduces the concept of classes and properties, allowing for the definition of vocabularies and a limited form of structure to be added to RDF data. For instance, using RDFS, it is possible to define hierarchies of classes and properties, specify the domain and range of properties, and declare subclasses or subproperties.
For example, the statement:
\small{\begin{verbatim}
    ex:Aventa_AV-7 rdf:type ex:WindTurbine
\end{verbatim}}%
\noindent
made in RDF can be semantically enriched by stating that Wind Trubine is a class.
\small{\begin{verbatim}
    @prefix ex: <http://example.com/resource/> .
    @prefix rdf: <http://www.w3.org/1999/02/22-rdf-syntax-ns#> .
    @prefix rdfs: <http://www.w3.org/2000/01/rdf-schema#> .
    
    ex:WindTurbine rdf:type rdfs:Class .  
    ex:Aventa_AV-7 rdf:type ex:WindTurbine .
    \end{verbatim}
}
\paragraph{Web Ontology Language (OWL)}
Web ontology language is a formal language based on the Description Logic representation formalism. Developed by the W3C, OWL is built on top of RDF and extends its expressiveness by providing additional vocabulary for defining complex relationships, classes, properties, and restrictions. OWL enables a higher level of semantic expressiveness compared to RDF and RDF Schema (RDFS), allowing for more sophisticated reasoning and inferencing capabilities.
For example we can express rated power as datatype property, and constrain it to be a float:
\small{
    \begin{verbatim}
    @prefix ex: <http://example.com/resource/> .
    @prefix rdf: <http://www.w3.org/1999/02/22-rdf-syntax-ns#> .
    @prefix rdfs: <http://www.w3.org/2000/01/rdf-schema#> .
    @prefix owl: <http://www.w3.org/2002/07/owl#> .
    @prefix xsd: <http://www.w3.org/2001/XMLSchema#> .
    
    ex:WindTurbine rdf:type owl:Class .
    ex:Aventa_AV-7 rdf:type ex:WindTurbine .
    ex:ratedPower rdf:type owl:DatatypeProperty ;
        rdfs:domain ex:WindTurbine ;
        rdfs:range xsd:float .
    ex:Aventa_AV-7 ex:ratedPower "6.2"^^xsd:float .
\end{verbatim}
}
There are several sublanguages of OWL with varying levels of expressiveness and computational complexity, including OWL Lite, OWL DL, and OWL Full. OWL DL, which is based on description logic, offers a balance between expressiveness and computational tractability, making it suitable for many applications.

\paragraph{Shapes Constraint Language (SHACL)}
Shapes Constraint Language (SHACL) is a World Wide Web Consortium (W3C) specification for validating and describing RDF graphs. SHACL allows for the definition of constraints that can be used to validate RDF data against a set of conditions. SHACL's validation capability makes it particularly suited for ensuring that data adheres to a particular shape or structure, hence the name. In addition to validation, SHACL can be used for data modelling and to guide the process of data transformation and integration. 
For example, the Aventa AV-7 wind turbine described before can be validated by a SHACL shape like:

\small{
\begin{verbatim}
    @prefix ex: <http://example.com/resource/> .
    @prefix rdf: <http://www.w3.org/1999/02/22-rdf-syntax-ns#> .
    @prefix sh: <http://www.w3.org/ns/shacl#> .
    @prefix xsd: <http://www.w3.org/2001/XMLSchema#> .
    
    ex:WindTurbineShape a sh:NodeShape ;
        sh:targetClass ex:WindTurbine ;
        sh:property [
            sh:path ex:ratedPower ;
            sh:datatype xsd:float ;
        ] .
\end{verbatim}
}
This SHACL shape defines that every instance of the class "WindTurbine" must have a property "ratedPower" which has to be of datatype float. This way, SHACL helps to ensure data integrity and consistency by providing a mechanism for enforcing data constraints.

\paragraph{JSON Schema}
JSON Schema represents a vocabulary permitting annotation and validation of JSON data. Unlike RDF, JSON stores information as attribute-value pairs. This type of data structure, when nested can be visualised as a tree rather than a labeled graph. JSON Schema, hence, defines the structure of JSON data and validates JSON data against defined schemas. It supports various constraints, such as data types, enumerations, pattern matching, optional/required properties, and array item uniqueness. 
For example, the information about Aventa AV-7 wind tubine can be stored in JSON format as following:
\begin{verbatim}
{
 "WindTurbine":{
  "model":"Aventa AV-7"
  "ratedPower":6.2
  }
}
\end{verbatim}
The related schema would look like:
\small{
\begin{verbatim}
{
    "$schema": "http://json-schema.org/draft-07/schema#",
    "title": "Wind Turbine",
    "description": "Schema for basic Wind Turbine attributes",
    "type": "object",
    "properties": {
        "WindTurbine": {
        "type": "object",
        "properties": {
            "model": {
                "type": "string",
                "description": "The model of the Wind Turbine"
            },
            "ratedPower": {
                "type": "number",
                "description": "The rated power of the Wind Turbine in kilowatts"
    }
    },
    "required": ["model", "ratedPower"]
    }
    },
    "required": ["WindTurbine"]
}
\end{verbatim}}
\noindent
Although primarily designed for JSON data validation, its use for more intricate data modeling tasks has been increasing, indicating its evolution towards a comprehensive ontology language.

\paragraph{YAML Schema}
YAML Schema is a tool dedicated to defining the structure of YAML documents. YAML, a human-friendly data serialization standard, is extensively used in configuration files and applications where data storage or transmission is involved. YAML Schema bears several similarities to JSON Schema, but it is designed specifically for the YAML data format. This schema validates YAML documents, ensuring compliance with a predefined structure and specific criteria. YAML and YAML Schema are commonly used when data and schema readability is paramount.
Below is a YAML data and YAML Schema serialisation of the same information as one described before by JSON and JSON Schema.

YAML Data:
\begin{verbatim}
    WindTurbine:
        model: Aventa AV-7
        ratedPower: 6.2
\end{verbatim}
\clearpage
Related schema:
\small{\begin{verbatim}
    $schema: http://json-schema.org/draft-07/schema#
    title: Wind Turbine
    description: Schema for basic Wind Turbine attributes
    type: object
    properties:
      WindTurbine:
        type: object
        properties:
          model:
            type: string
            description: The model of the Wind Turbine
          ratedPower:
            type: number
            description: The rated power of the Wind Turbine in kilowatts
        required: 
          - model
          - ratedPower
    required:
      - WindTurbine 
\end{verbatim}}

\paragraph{XML Schema}
XML Schema, also known as XSD (XML Schema Definition), is a World Wide Web Consortium (W3C) recommendation that prescribes formal descriptions of elements in an Extensible Markup Language (XML) document. It serves to describe and validate the structure and content of XML data. XML Schema supports namespaces, complex data types, inheritance (through extension and restriction), and constraints on values and relationships between elements. Its wide use spans several industries, including publishing, telecommunications, and e-commerce. However due complexity of XML syntax, this schema languages is much less intuitive and is not as easily humanly readable as compared to JSON or YAML Schema.
\small{\begin{verbatim}
<?xml version="1.0"?>
<xs:schema xmlns:xs="http://www.w3.org/2001/XMLSchema">
  <xs:element name="WindTurbine">
    <xs:annotation>
      <xs:documentation>A representation of a Wind Turbine.</xs:documentation>
    </xs:annotation>
    <xs:complexType>
      <xs:attribute name="model" type="xs:string">
        <xs:annotation>
          <xs:documentation>The model of the Wind Turbine.</xs:documentation>
        </xs:annotation>
      </xs:attribute>
      <xs:attribute name="ratedPower" type="xs:float">
        <xs:annotation>
          <xs:documentation>The rated power of the Wind Turbine in megawatts.</xs:documentation>
        </xs:annotation>
      </xs:attribute>
    </xs:complexType>
  </xs:element>
</xs:schema>
\end{verbatim}}


\section{Serialisation formats}
\label{app:serialisation}
OWL and RDFS ontologies can be serialised for storage using a variety of formats. These formats enable the representation of knowledge in a machine-readable and standardized way. RDF, as the foundation of the Semantic Web, can be serialized in various formats such as RDF/XML, Turtle (Terse RDF Triple Language), N-Triples, and JSON-LD. 

RDF/XML was the first standardized serialisation format for RDF, but its verbosity and complexity led to the development of other formats such as Turtle and N-Triples, which offer more human-readable syntax. JSON-LD (JSON for Linked Data) has gained popularity as it combines the simplicity and widespread use of JSON with the ability to express RDF data.
OWL, being an extension of RDF, can also be serialised using the aforementioned RDF serialisation formats. However, OWL has its own serialisation formats as well, such as OWL/XML and Functional-Style Syntax (also known as OWL Functional Syntax). OWL/XML is an XML-based syntax specifically designed for expressing OWL ontologies, while Functional-Style Syntax is a human-readable, text-based format that closely follows the structure of the OWL 2 specification. The choice of serialization format depends on factors such as readability, compatibility with existing tools, and ease of parsing and processing.

\paragraph{JavaScript Object Notation for Linked Data (JSON-LD) }
JSON-LD is a lightweight data interchange format that extends JSON to provide a means for encoding Linked Data using standard JSON conventions. JSON-LD is designed to be easy to read and write by humans, as well as simple to parse and generate by machines. It is developed by the World Wide Web Consortium (W3C) and provides a way to represent RDF data model in JSON.   
JSON-LD is particularly useful for web developers who want to incorporate structured data into web applications and APIs while leveraging the existing JSON tools and libraries.
JSON-LD introduces a notion of a context, which allows defining short aliases for long IRIs (Internationalized Resource Identifiers) used in RDF, simplifying the representation of RDF triples in JSON. It also supports the definition of data types, language tags for string values, and nested JSON objects to represent complex relationships and structures.
JSON-LD serialisation example:

\begin{verbatim}
{
  "@context": {
    "schema": "https://schema.org/",
    "geo": "http://www.w3.org/2003/01/geo/wgs84_pos#"
  },
  "@type": "schema:EducationalOrganization",
  "@id": "https://www.ost.ch",
  "schema:name": "OST - Ostschweizer Fachhochschule",
  "schema:owns": {
    "@type": "schema:Product",
    "@id": "urn:ost:aventa:av-7",
    "schema:name": "Aventa AV-7 Wind Turbine",
    "schema:manufacturer": {
      "@type": "schema:Organization",
      "@id": "https://en.wind-turbine-models.com/manufacturers/316-aventa",
      "schema:name": "Aventa"
    },
    "schema:model": {
      "@type": "schema:ProductModel",
      "@id": "https://en.wind-turbine-models.com/turbines/1529-aventa-av-7",
      "name":"AV-7"
    },
    "geo:location": {
      "@type": "geo:Point",
      "geo:lat": "47.52000",
      "geo:long": "8.68236"
    }
  }
}
\end{verbatim}

\begin{figure}[b]
    \centering
    \includegraphics[width=\textwidth]{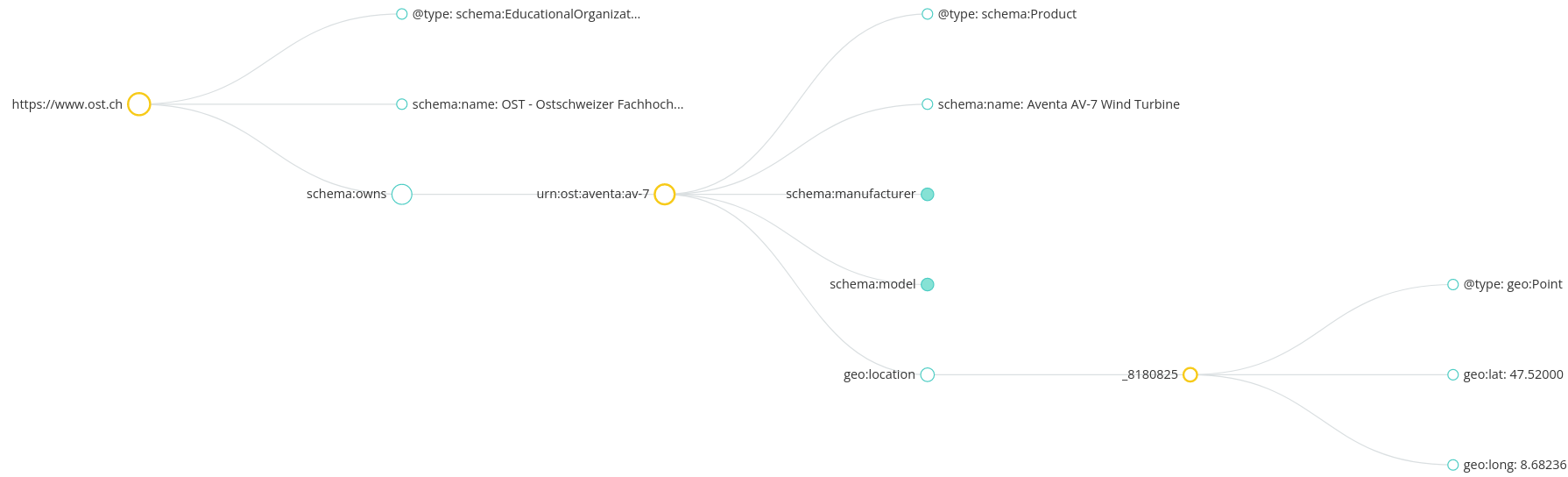}
    
    \caption{Aventa AV-7 described using JSON-LD}
    \label{fig:aventa-json-ld}
\end{figure}

\clearpage
\section{FAIR Principles and Semantic Web Technologies}
\label{app:fair_ld}

\begin{table}[h]
\begin{tabular}{p{2cm} p{13.5cm}}
\parbox[t]{2cm}{
\textbf{FAIR} \\
\textbf{Principle}
} & \textbf{Semantic Web Technology Enablement} \\
\hline \hline
Findable &\\
\hline
F1 & URIs, especially HTTP URIs are globally unique and can be persistent.  This requirement is similar to the first LD  principle, even though persistence is not specifically addressed by LD.\\
F2 & Rich metadata models are developed and adopted by communities, for their specific domain. These models can be published on the web as LD and referred to, when describing the data. \\
F3 & RDF triples connect unique identifiers (URIs) with well-defined properties to their respective values. \\
F4 & RDF data can be stored in Triplestores which are optimized for semantic queries using SPARQL. This requirement along with F2 and F3 are similar to the third LD principle. \\
\hline
Accessible&\\
\hline
A1.1 & Semantic web technologies are built on open standards such as HTTP and RDF. This requirement is simmilar to the second linked data principle.\\
A1.2 & Web protocols like HTTPS include provisions for authentication and authorization. \\
A2 & Persistent URIs and versioning methodologies ensure that metadata remain accessible. Metadata models can be published on third party websites like schema.org, more expressive semantic artefacts can be published on ontology hosting websites like the ones powered by OntoPortal. This requirement has no equivalent in LD principles\\
\hline 
Interoperable&\\
\hline
I1 & RDF, RDFS, and OWL are formal languages that are broadly applicable for knowledge representation. \\
I2 & Semantic web enables the use of vocabularies and taxonomies through standard data models such as SKOS.\\
I3 & RDF's linking capability enables metadata to include qualified references to other metadata. This requirement is similar to the fourth LD principle. \\
\hline 
Reusable&\\
\hline
R1.1 & Metadata in RDF can include licensing information using appropriate vocabularies. In general, LD has evolved into the direction of open data, meanwhile FAIR principles can applied to data subject to any explicitly defined license.\\
R1.2 & Provenance information can be recorded using cross-domain, community accepted vocabularies such as PROV-O. \\
R1.3 & The semantic web supports the use of domain-specific ontologies, which can be developed according to community standards. This requirement is similar to the third LD principle.\\
\hline
\end{tabular}
\caption{FAIR Principles and Semantic Web Technology Enablement}
\label{table:fair_semantic_web}
\end{table}

\bibliographystyle{unsrt}
\bibliography{references}

\end{document}